\documentclass{article} 
\usepackage{iclr2027_conference,times}

\usepackage{amsmath,amsfonts,bm}

\def\eqref#1{equation~\ref{#1}}

\def\1{\bm{1}}

\DeclareMathAlphabet{\mathsfit}{\encodingdefault}{\sfdefault}{m}{sl}
\SetMathAlphabet{\mathsfit}{bold}{\encodingdefault}{\sfdefault}{bx}{n}

\usepackage{adjustbox}
\usepackage{multirow}
\usepackage{booktabs}
\usepackage{wrapfig}
\usepackage{url}
\usepackage[dvipsnames]{xcolor}
\usepackage[
    colorlinks=true,
    urlcolor=RoyalBlue,
    citecolor=RoyalBlue
]{hyperref}

\newcommand{\ours}{DIGHT}

\title{From Digital Human Interactions to Physics-Based Humanoid Skills: Physics-Grounded Post-Training of Interaction Generators}

\author{Kerui Chen$^{1}$, Jianrong Zhang$^{1}$, Kai Lv$^{2}$, \& Hehe Fan$^{1*}$ \\
$^{1}$College of Artificial Intelligence, Zhejiang University. \\ 
$^{2}$School of Computer Science and Technology, Beijing Jiaotong University. \\
*Corresponding author. \\
}

\begin{document}

\maketitle

\begin{abstract}
Recent methods have made promising progress in generating interactions between two humanoids, largely relying on physics-based tracking policies to convert digital reference motions into executable trajectories. However, limited tracking capabilities restrict the range of reference motions that can be successfully executed, reducing data utilization. Moreover, even successful tracking does not guarantee physically plausible responses or faithful realization of the intended interactions. In this paper, we introduce DIGHT, a co-adaptive framework that couples a \textbf{D}igital human \textbf{I}nteraction \textbf{G}enerator with a \textbf{H}umanoid \textbf{T}racking policy. 
Our DIGHT first executes multiple text-conditioned interaction candidates in simulation using a fixed tracker. It then constructs physics-grounded preferences from the resulting rollouts, covering both general executability and interaction fidelity. 
Rather than collapsing these signals into a single scalar reward for candidate ranking, we align the pretrained generator using physics-decoupled diffusion direct preference optimization (DPO), preserving criterion-specific supervision without differentiating through the simulator. To improve executability, preference pairs are derived from tracking error, friction, and floating. Additionally, to improve interaction fidelity, we propose to incorporate force feedback from the simulator as a measure of contact fidelity and construct preferences over contact occurrence, location, duration, and force magnitude. The aligned generator then supplies reference motions for fine-tuning the tracker, improving compatibility between generation and physical execution. Extensive experiments demonstrate that our approach not only improves the physical plausibility of generated motions but also enables more reliable and faithful humanoid interactions in simulation.
Project page: \url{https://1233chen.github.io/DIGHT/}.
\end{abstract}
\section{Introduction}
Physics-based interaction between two humanoids is an important capability for embodied agents, enabling them to coordinate with and physically respond to one another in a shared environment. It supports a wide range of applications, including collaborative agents, social robotics, character animation, and embodied simulation~\citep{sui2025survey, vall2024robot, zhang2023simulation}. In particular, generating such interactions from textual descriptions provides an intuitive way to specify diverse behaviors~\citep{li2026interagent, javed2024intermask,wang2025timotion} while requiring the resulting motions to be both semantically meaningful and physically realizable.

Relative to single-humanoid control, dual-humanoid coordination introduces heightened complexities~\citep{zhang2026contactrich}. In addition to ensuring individual stability and dynamic feasibility, the agents must synchronize their spatial configurations, motion timing, and physical interactions~\citep{zhang2023simulation, liu2026hint, lionar2026teamhoi}. Moreover, their executions are inherently coupled: the motion and contact forces of one humanoid directly affect the state and response of the other~\citep{assistmimic}. Consequently, even small deviations in position or timing can lead to missed contacts, implausible forces, or unstable executions, ultimately altering the intended interaction~\citep{huang2026humanhumanoid, multiphys2024}. Therefore, successful two-humanoid interaction requires not only \textit{general executability} for each participant, but also \textit{interaction fidelity} that preserves the intended contacts and coordination between them.

Recently, many works have achieved humanoid interaction generation by leveraging humanoid tracking policies in two ways. Human-X~\citep{human-x} first generates digital human motions and then executes them using a humanoid tracking policy for reaction generation. In contrast, PhysReaction~\citep{liu2024physreaction} and InterAgent~\citep{li2026interagent} use tracking policies to convert MoCap data into executable humanoid trajectories, which serve as supervision for training interaction generators.

\begin{figure}
    \centering
    \includegraphics[width=0.95\linewidth]{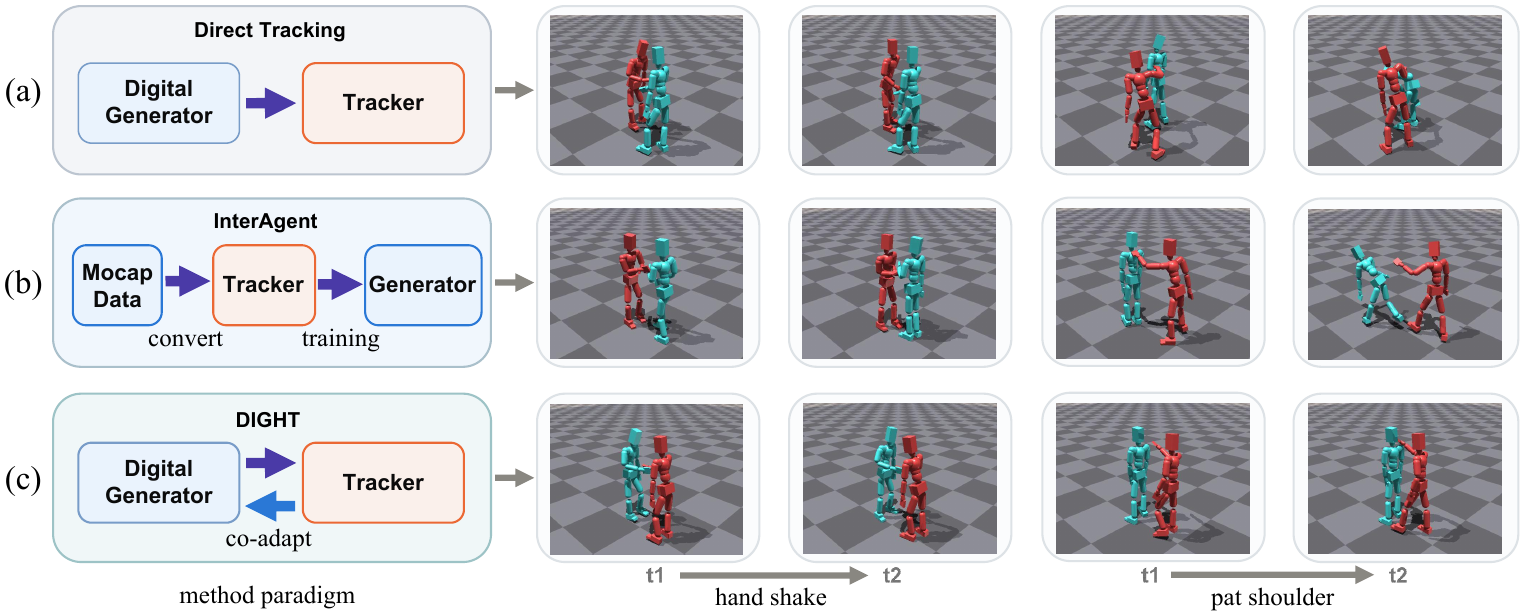}
    \vspace{-5mm}
    \caption{\textbf{Comparison between existing methods and \ours.} (a) Direct tracking may miss intended contacts or induce excessive forces despite successful pose tracking. (b) InterAgent produces physically implausible contact forces and unstable interactions due to the gap between the training and inference. (c) In contrast, \ours~uses physical feedback from generated rollouts to co-adapt the interaction generator and tracking policy, preserving intended contacts while producing physically stable interactions.}
    \vspace{-6mm}
    \label{fig:intro}
\end{figure}

However, both paradigms inherit fundamental limitations from their reliance solely on tracking.
As shown in Fig.~\ref{fig:intro}(a), direct tracking primarily optimizes pose-level similarity with a generated reference rather than the physical fidelity of the interaction. Consequently, even a successfully tracked interaction may exhibit unrealistic contact behavior: the handshake may lose the intended hand-to-hand contact, while following the shoulder-patting reference may induce excessive contact forces that destabilize the receiving humanoid.
On the other side, as shown in Fig.~\ref{fig:intro}(b), although InterAgent trains its generator on trajectories obtained through humanoid tracking, it learns physical feasibility only indirectly by imitating these trajectories. This leaves a gap between training and generation, as the generated interactions are not subject to explicit physics-based constraints on their dynamics at inference time. Consequently, the generated patting shoulder interaction produces a physically implausible contact force that destabilizes the receiving humanoid.

In this paper, we introduce \textbf{\ours}, a physics-grounded co-adaptive framework that couples a \textbf{D}igital human \textbf{I}nteraction \textbf{G}enerator with a \textbf{H}umanoid \textbf{T}racking policy. The core idea behind \ours~is to bridge digital generation and physical execution through physics-grounded co-adaptation: physical feedback from tracking rollouts informs the generator about the physical capabilities of target humanoids, while reference motions from the refined generator are used to adapt the tracking policy to the generator's motion distribution. By enabling each component to account for the other's capabilities, \ours~narrows the gap between digital interaction generation and physical humanoid execution. Specifically, \ours~samples multiple candidate interactions and executes them in simulation via the tracking policy. It then translates their relative performance in executability and interaction fidelity into criterion-specific preferences to optimize the generator. For improving executability, we use three general physical preferences including floating, friction, and tracking errors. To address missing and unrealistic contacts in humanoid interaction, we propose to incorporate force feedback from simulator to construct contact-related preferences that assess contact occurrence and interaction fidelity across contact locations, durations, and force magnitudes. To preserve distinct supervision signals from different physical preferences, we introduce physics-decoupled diffusion DPO rather than aggregating them into a single scalar reward. Through this optimization, the generator is progressively refined toward motions that are physically executable and faithful to intended contacts. Finally, the aligned generator provides reference motions for policy adaptation, establishing a mutually reinforcing co-adapted framework.

We evaluate \ours~on two text-conditioned interaction datasets, InterHuman~\citep{liang2024intergen} and Inter-X~\citep{xu2024inter}, using two distinct diffusion backbones, InterGen~\citep{liang2024intergen} and TIMotion~\citep{wang2025timotion}. Across all four dataset-backbone settings, \ours~consistently reduces all five evaluated physical errors while largely preserving the motion quality and text alignment of the pretrained generators. Existing interaction datasets provide paired motions and textual descriptions, but lack explicit annotations of the contacts required by each interaction, making it difficult to directly evaluate whether a generated motion realizes the contacts implied by its text. To address this limitation, we establish a contact-aware benchmark based on the InterHuman test set, labeling whether an interaction requires physical contact and identifying the expected contacting body-region pairs. On this benchmark, \ours~improves contact-topology precision from 38.29\% to 46.77\% and recall from 52.16\% to 63.94\% over the InterGen baseline. In addition, our method generalizes effectively to temporal compositional and multi-role interaction scenarios.

Our contributions can be summarized as follows: \textbf{1.} We introduce \ours, a physics-grounded co-adaptive framework that jointly adapts interaction generation and humanoid tracking through physical execution feedback. \textbf{2.} We propose physics-decoupled diffusion DPO to align the interaction generator with multiple physics-grounded preferences without collapsing all signals into a single scalar reward. \textbf{3.} We explicitly incorporate interaction forces to constructing contact-related preferences that evaluate contact occurrence and interaction fidelity. \textbf{4.} We develop a contact-annotated evaluation subset of InterHuman to directly measure the fidelity of intended physical contacts.

\section{Related Work}

\noindent\textbf{Text-conditioned Human-human Interaction Generation.}
Text-conditioned human--human interaction generation aims to synthesize coordinated motions of two interacting individuals from textual descriptions. Early diffusion-based approaches such as InterGen~\citep{liang2024intergen} jointly generate two motion sequences and model cross-participant dependencies through mutual attention. Subsequent methods improve interaction modeling through participant specific language conditions~\citep{ruiz2024in2in}, collaborative masked prediction~\citep{javed2024intermask}, unified multi-person latent representations~\citep{li2025two}, temporal and role-evolving modeling~\citep{wang2025timotion}, and efficient architectures~\citep{wu2025intermamba}. 
Other works support fine-grained control~\citep{wu2026text2interact,wang2024intercontrol}, autoregressive interaction generation~\citep{ruiz2026interact2ar}, or generation for arbitrary numbers of participants~\citep{fan2024freemotion, yu2026socialgen}. 
More recent approaches introduce physical priors into interaction generation. PhyInter~\citep{gao2025physics} incorporates physics-guided constraints into motion diffusion, while PhysiInter~\citep{yao2025physiinter} uses physical mapping to improve the fidelity of generated interactions. 
These works optimize their models primarily in kinematic motion space, without considering the physical realism of generated interactions when executed in a physics environment. In this work, we target on physically realistic humanoid interaction execution while simultaneously improving the kinematic interactions.

\noindent\textbf{Physics-based Humanoid Interaction.}
Physics-based humanoid control learns policies that reproduce reference motions while satisfying embodiment and environmental dynamics. DeepMimic~\citep{peng2018deepmimic} established reinforcement learning-based motion imitation for simulated characters. Subsequent methods learned reusable motion priors and skill representations from larger motion collections, including AMP~\citep{peng2021amp}, ASE~\citep{peng2022ase}, PHC~\citep{luo2023perpetual}, PULSE~\citep{PULSE}, and MoConVQ~\citep{yao2024moconvq}. These advances enabled robust tracking and control across diverse motions rather than individual reference clips. 
Recent research has further extended physics-based control from individual characters to contact-rich interactions, developing interaction-aware objectives to preserve spatial coordination, contact relationships, and coupled dynamics in human-object and multi-character settings~\citep{zhang2023simulation,wang2023physhoi,multiphys2024, xu2025intermimic, wang2025skillmimic, assistmimic}. 

Building on advances in humanoid control, some studies achieve conditioned humanoid generation in two paradigms. CLoSD~\citep{tevet2025closd}and Human-X~\citep{human-x} first generate digital interaction motion, and then execute it using a humanoid tracking policy. In contrast, Uniphys~\citep{wu2025uniphys}, PhysReaction~\citep{liu2024physreaction} and InterAgent~\citep{li2026interagent} use trajectories obtained through humanoid tracking as supervision for training humanoid generation models.
These works demonstrate that physically humanoid execution can be realized by leveraging the capability of tracking policy. However, in complex two-humanoid interactions, relying solely on the tracking policy can lead to missing intended contacts and physically implausible interaction forces. To address this challenge, \ours~introduces a co-adaptive framework that couples the interaction generator with the tracking policy through physical execution feedback and explicitly incorporates interaction forces into preference construction for generator alignment.

\noindent\textbf{Physical Feedback for Motion Model Alignment.} 
Physical feedback has recently been used to improve the executability of generated human motion. PhysDiff~\citep{yuan2023physdiff} and CLoSD~\citep{tevet2025closd} incorporate physics-based refinement or control during inference.
RLPF~\citep{yue2025rl} and PhysMoDPO~\citep{zhang2026physmodpo} instead optimize a motion generator using the feedback from a frozen humanoid tracking policy, making the generated motion more physical plausible.
In this work, we incorporate force feedback from the simulator to address physically unrealistic contacts in two-humanoid interactions.

\section{Preliminaries}
\textbf{Interaction Tracking Policy.}
Following the previous works, we adopt reinforcement learning to learn a tracking policy for imitating motion capture data. Similar to PHC, we utilize a full SMPL humanoid controlled by PD controllers. The learned policy outputs target joint rotations as actions $a_t=\pi(s_t,s_{t+1}^{ref})$, where $s_t$ and $s_{t+1}^{ref}$ denote the current agent state and the future reference state, respectively. Beyond the tracking rewards used in PHC, we adopt an interaction graph reward~\citep{zhang2023simulation, li2026interagent} to explicitly model the spatial relationships between agents, encouraging the policy to preserve the inter-agent configurations observed in the reference motions.

\textbf{Data Retarget.}
To bridge the morphology gap between the tracking policy and the digital generator, we retarget the motion data to a unified humanoid morphology used in the simulation environment. Since retargeting may introduce contact misalignment, such as palm contacts shifting to the wrist or forearm~\citep{araujo2025retargeting}, we further apply geometry-based contact refinement to recover the original contact relationships while preserving the overall motion structure. The detailed retargeting algorithm is provided in the Appendix~\ref{app:motion-retarget}.

\textbf{Interaction Diffusion Model.}
Text-conditioned human-human interaction models learn the distribution of coordinated motions from MoCap data.
Given a clean two-person motion $x_0=(x_0^A,x_0^B)$ and text description $c$, the forward diffusion process produces
$q(x_t\mid x_0)
=
\mathcal{N}\!\left(
\sqrt{\bar{\alpha}_t}\,x_0,
(1-\bar{\alpha}_t)I
\right)$,
where $t$ is the diffusion timestep and $\bar{\alpha}_t$ is determined by a predefined noise schedule. The denoiser jointly processes the two participants and predicts the clean interaction,
$
\hat{x}_0
=
G_{\theta}(x_t^A,x_t^B,t,c).
$
Iterative reverse diffusion then defines the conditional motion distribution $p_{\theta}(x_0\mid c)$. We use InterGen~\citep{liang2024intergen} and TIMotion~\citep{wang2025timotion} as generator backbones, and both are optimized in the clean-motion prediction parameterization used by our preference objective.

\section{Method}
 
\subsection{Overview}
\begin{figure}[t]
    \centering
    \includegraphics[width=0.95\linewidth]{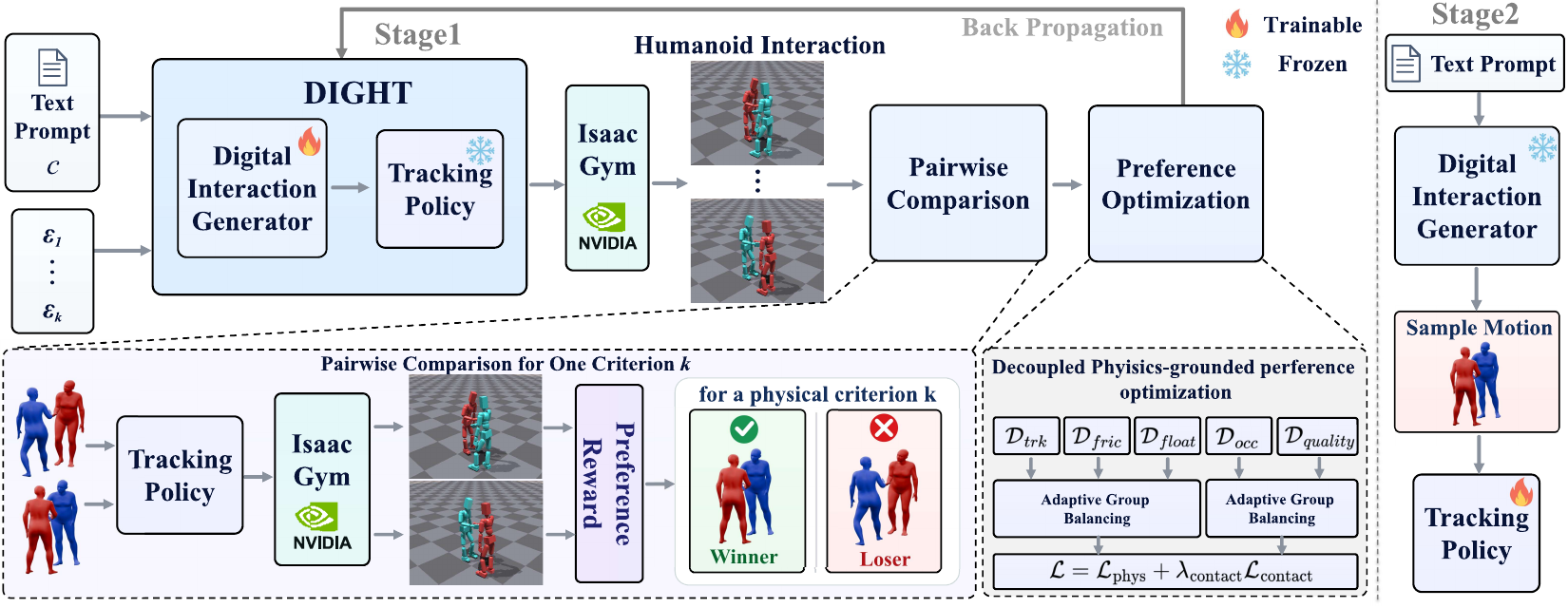}
    \vspace{-3mm}
    \caption{\textbf{Overview of \ours.} \ours~consists of an interaction generator and a tracking policy, which are co-adapted in two stages. First, multiple interaction candidates produced by the generator are executed by the frozen tracker, whose rollouts are used to construct physics-grounded preferences for tracking, frictional dissipation, floating, contact occurrence, and contact quality. These preferences align the generator through physics-decoupled diffusion DPO. Second, motions from the aligned generator are used to adapt the tracking policy, improving their compatibility.}
    \label{fig:method}
    \vspace{-4mm}
\end{figure}

\ours~couples a digital interaction generator $g_\theta(x\mid c)$ with a physics-based humanoid tracking policy $\pi_\phi$. Given text $c$, the generator generates a digital interaction motion $x=(x^A,x^B)$, and the tracker executes it in simulation to produce a physics-based humanoid interaction motion $\tau=\operatorname{Sim}(x;\pi_\phi)$.
As shown in Fig.~\ref{fig:method}, \ours~performs generator-tracker co-adaptation in two stages. In the first stage, the pretrained generator samples multiple candidate motions for each training prompt. The frozen pretrained tracker executes these candidates in simulation. The resulting rollouts are then evaluated to construct five physics-grounded preference sets based on tracking error, frictional dissipation, floating, contact occurrence, and contact quality, which are used to post-train the generator. Second, we freeze the aligned generator and use its sampled motions as references to adapt the tracker to its original tracking objective. The two components are therefore adapted using information induced by each other.

\subsection{Physics-Grounded Preference Construction}
For each text-motion example $(c,x^{\mathrm{gt}})$, we sample $K$ candidates $x_i\sim g_\theta(x\mid c)$ and execute them with the same frozen tracker, obtaining $\tau_i=\operatorname{Sim}(x_i;\pi_\phi)$. We also execute $x^{\mathrm{gt}}$ to obtain $\tau^{\mathrm{gt}}$, which provides the reference contact pattern but is not automatically used as the preferred sample. We construct five physics-grounded preference pairs as follows.

\textbf{General Physical Preferences.}
For every rollout, we compute three general physical scores: mean body tracking deviation $E_{\mathrm{trk}}$, frictional dissipation $E_{\mathrm{fric}}$, and the float $E_{\mathrm{float}}$. For each criterion $k\in\{\mathrm{trk},\mathrm{fric},\mathrm{float}\}$, a valid winner must have a lower corresponding error than the loser:
\begin{equation}
    E_k(\tau^+) < E_k(\tau^-),
    \qquad
    k\in\{\mathrm{trk},\mathrm{fric},\mathrm{float}\},
    \label{eq:general_physical_preference}
\end{equation}
where $\tau^+$ and $\tau^-$ are the executions of the preferred and rejected candidates, respectively. The winner is additionally required to have better motion-text alignment than the loser.

\noindent\textbf{Contact Occurrence Preference.}
Beyond general physical plausibility, two-person interactions differ from single-person in their need to preserve meaningful contact. To capture such contact, we leverage the force feedback provided by the physics simulator. We detect frame-level contacts between rigid bodies and then consolidate them into persistent contacts between semantic body regions. Specifically, each rigid body is assigned to one of six regions: the core, head, arms, hands, legs, or feet. At each frame, inter-person contact forces are grouped according to the corresponding pair of body regions. For every region pair, we slide a fixed-length temporal window over the sequence and compute the contact ratio, defined as the proportion of frames within the window in which contact forces are detected. A region pair is labeled as a persistent contact if its contact ratio reaches a predefined threshold in at least one window.
This procedure filters out isolated collisions while remaining tolerant to brief interruptions in otherwise sustained contacts. We collect all persistent region pairs in $\mathcal{C}(\tau)$, which denotes the persistent contact topology of trajectory $\tau$. We use this topology to construct contact occurrence preferences. If $\mathcal{C}(\tau^{\mathrm{gt}})$ is nonempty, the winner must reproduce at least one reference contact without introducing any persistent contact outside $\mathcal{C}(\tau^{\mathrm{gt}})$, whereas the loser must contain no persistent contact. If $\mathcal{C}(\tau^{\mathrm{gt}})$ is empty, the winner must remain contact-free, whereas the loser must contain at least one persistent contact.

\noindent\textbf{Contact Quality Preference.}
For contact-positive motions, contact occurrence alone does not ensure interaction quality: the contact may be transient or poorly sustained, and even persistent contact may involve implausibly weak or excessive forces. We therefore measure three complementary aspects of contact quality relative to the $\tau^{\mathrm{gt}}$: $E_{\mathrm{topo}}$ measures the topology fidelity of persistent region-level contacts, $E_{\mathrm{dur}}$ measures the duration fidelity of the longest episode for each corresponding contact edge, and $E_{\mathrm{force}}$ measures its force fidelity in terms of the mean and maximum forces within that episode. A contact quality winner must be better in all component:
\begin{equation}
  E_j(\tau^+)< E_j(\tau^-),\ \forall j,
  \label{eq:contact_quality_preference}
\end{equation}
where $j\in\{\mathrm{topo},\mathrm{dur},\mathrm{force}\}$. These pairs are also constrained by the motion-text alignment.
Finally, we construct the five independent preference sets
$\mathcal{D}_k=\{(c,x^+,x^-):x^+\succ_k x^-\}$ for
$k\in\mathcal{T}=\{\mathrm{trk},\mathrm{fric},\mathrm{float},\mathrm{occ},\mathrm{quality}\}$.
The detailed definitions of each preference score are provided in Appendix~\ref{app:method-details}.

\subsection{Physics-Decoupled Diffusion DPO}
A motion can be physically preferable under one criterion while being inferior under another. Collapsing all physical signals into a single scalar reward for candidate ranking, discarding criterion-specific supervision. We therefore optimize a separate decoupled diffusion DPO objective for each preference set $\mathcal{D}_k$. During generator alignment, the trainable denoiser $\theta$ and the frozen reference denoiser $\theta_{\mathrm{ref}}$ are initialized from the same pretrained generator. Since both backbones predict the clean motion, we define the relative denoising reward
\begin{equation}
    r_\theta(x,t,c)
    =
    \left\|x-\hat{x}_{\theta_{\mathrm{ref}}}(x_t,t,c)\right\|_2^2
    -
    \left\|x-\hat{x}_{\theta}(x_t,t,c)\right\|_2^2,
    \label{eq:relative_denoising_reward}
\end{equation}
where $x_t$ is the noisy motion at timestep $t$ and $\hat{x}_\theta$ denotes the predicted clean motion. For a preference pair $(c,x^+,x^-)\sim\mathcal{D}_k$, we define
\begin{equation}
    R_k
    =
    r_\theta(x^+,t,c)-r_\theta(x^-,t,c),
    \qquad
    \mathcal{L}_k
    =
    -\mathbb{E}
    \left[
      \log\sigma\!\left(\beta R_k\right)
    \right],
    \label{eq:physical_dpo}
\end{equation}
where the expectation is over preference pairs, uniformly sampled diffusion timesteps, and independent forward-process noise for $x^+$ and $x^-$. Here, $\sigma$ is the sigmoid function and $\beta>0$ controls the preference strength. A larger $R_k$ indicates that the trainable denoiser improves the preferred motion over the rejected one relative to the frozen reference denoiser.

We group tracking, friction, and floating into $\mathcal{G}_{\mathrm{phys}}$, and occurrence and contact quality into $\mathcal{G}_{\mathrm{contact}}$. Following BideDPO~\citep{zhou2026bidedpo}, for either group $\mathcal{G}\in\{\mathcal{G}_{\mathrm{phys}},\mathcal{G}_{\mathrm{contact}}\}$, their current mini-batch losses determine detached within-group weights:
\begin{equation}
    \alpha_k^{(\mathcal{G})}
    =
    \operatorname{sg}\!\left(
      \frac{\mathcal{L}_k}
      {\sum_{j\in\mathcal{G}}\mathcal{L}_j}
    \right),
    \qquad
    \mathcal{L}_{\mathcal{G}}
    =
    \sum_{k\in\mathcal{G}}
    \alpha_k^{(\mathcal{G})}\mathcal{L}_k,
    \qquad
    \mathcal{L}
    =
    \mathcal{L}_{\mathcal{G}_{\mathrm{phys}}}
    +
    \lambda_{\mathrm{contact}}
    \mathcal{L}_{\mathcal{G}_{\mathrm{contact}}},
    \label{eq:contact5_objective}
\end{equation}
where $\operatorname{sg}$ denotes stop-gradient and $\lambda_{\mathrm{contact}}$ balances contact fidelity against general executability. Only the trainable denoiser $\theta$ is updated.

\subsection{Tracker Adaptation}
After aligning the generator, we use it to provide reference motions for fine-tuning the tracker. Specifically, we freeze the aligned generator $g_{\theta_{\mathrm{align}}}$, sample interaction motions $x\sim g_{\theta_{\mathrm{align}}}(x\mid c)$, and use them as the reference motions to fine-tune the pretrained tracker with its original tracking objective. This stage adapts the tracker to the motion distribution of the aligned generator, thereby further improving compatibility between motion generation and physical execution. At inference time, $p_{\theta_{\mathrm{align}}}$ generates the reference interaction and the adapted tracker $\pi_{\phi_{\mathrm{adapt}}}$ executes it. 

\section{Experiments}

\subsection{Experiment Setup}
In our experiments, we apply our approach to the commonly used model InterGen~\citep{liang2024intergen} and the state-of-the-art model TIMotion~\citep{wang2025timotion}. We evaluate both the quality of their generated digital motions and the quality of their physical execution by humanoid agents.

\textbf{Datasets and Evaluation Configurations.}
We conduct experiments on two datasets, InterHuman~\citep{liang2024intergen} and Inter-X~\citep{xu2024inter} datasets, which provide text annotations and are designed for evaluating multi-human interaction generation. Our evaluation has two parts: generator evaluation assesses the quality of the generated digital interaction motions, while humanoid execution evaluation assesses their physical execution by the target humanoid tracker in the simulation environment. The results of Inter-X are shown in Appendix~\ref{app:inter_x}.

\begin{table*}[t!]
\centering
\caption{\textbf{Quantitative comparisons on InterHuman test set.} The best results are \textbf{bold}. $\text{\ours}^{1}$ and $\text{\ours}^{2}$ denote our framework initialized from InterGen and TIMotion, respectively. The Force Error is evaluated from simulated executions using the same pretrained tracker, while the other metrics evaluate the outputs of the generators.}
\label{tab:main_interhuman}
\begin{adjustbox}{width=0.95\linewidth}
\begin{tabular}{l|ccc|ccccc}
\toprule
\multirow{2}{*}{Methods} & \multirow{2}{*}{FID\ $\downarrow$}& \multirow{2}{*}{MM Dist\ $\downarrow$} & {R-Precision} & \multirow{2}{*}{Skate\ $\downarrow$} & \multirow{2}{*}{Float $\downarrow$} & \multirow{2}{*}{Interpen $\downarrow$} & \multirow{2}{*}{Jerk $\downarrow$} & \multirow{2}{*}{Force Error $\downarrow$}\\
&&& (Top-1) \ $\uparrow$&&&&& \\
\midrule
InterGen& 1.539 & 0.954 & 0.474 & 7.27 & 22.40 & 10.69 & 148.96 & 577.41 \\
InterGen + SFT & 1.878 & 0.965 & 0.461 & 6.57 & 21.54 & 13.12 & 129.13 & 543.62 \\
InterGen + DPO & 1.501 & 0.955 & 0.472 & 6.04 & 20.96 & 9.85 & 134.82 & 491.40 \\
$\text{\ours}^{1}$& \textbf{1.474} & \textbf{0.942} & \textbf{0.486} & \textbf{5.67} & \textbf{18.58} & \textbf{8.42} & \textbf{111.94} & \textbf{427.98} \\
\midrule
TIMotion& 1.341 & 0.931 & 0.536 & 11.69 & 21.63 & 9.68 & 322.02 & 619.25 \\
$\text{\ours}^{2}$& \textbf{1.297} & \textbf{0.927} & \textbf{0.543} & \textbf{8.53} & \textbf{18.28} & \textbf{7.83} & \textbf{199.76} & \textbf{451.33} \\
\bottomrule
\end{tabular}
\end{adjustbox}
\vspace{-5.8mm}
\end{table*}

\begin{table*}[t]
\centering
\caption{\textbf{Compared with humanoid generation on InterHuman test set.} The execution trajectory of \ours~is executed by the adapted tracker.}
\label{tab:interagent}
\begin{adjustbox}{width=0.95\linewidth}
\footnotesize
\begin{tabular}{l|ccccc|ccc}
\toprule
\multirow{2}{*}{Methods} 
& \multirow{2}{*}{FID $\downarrow$}
& \multirow{2}{*}{MM Dist $\downarrow$}
& \multicolumn{3}{c|}{R-Precision $\uparrow$}
& \multirow{2}{*}{Skate $\downarrow$}
& \multirow{2}{*}{Jerk $\downarrow$}
& \multirow{2}{*}{Force Error $\downarrow$} \\
&&&\multicolumn{3}{c|}{}&&& \\[-12pt]\cmidrule(lr){4-6}
&&&Top-1 & Top-2 & Top-3&&& \\
\midrule
InterAgent & 2.533 & 1.126 & 0.335 & 0.470 & 0.552 & 9.55 & 389.51 & 588.63 \\
$\text{\ours}$ & \textbf{1.426} & \textbf{1.054} & \textbf{0.417} & \textbf{0.565} & \textbf{0.644} & \textbf{6.23} & \textbf{231.28} & \textbf{413.47}\\
\bottomrule
\end{tabular}
\end{adjustbox}
\vspace{-4mm}
\end{table*}

\textbf{Evaluation Metrics.}
To comprehensively evaluate both the quality of generated interactions and their alignment with text descriptions, we adopt eight metrics: (1) R-Precision and MM-Dist for text-motion consistency; (2) FID for motion quality; (3) Float, Skate, Interpen, and Jerk for physical plausibility, including stability, foot contact, body penetration, and motion smoothness; and (4) Force Error for the interaction fidelity of humanoid execution by comparing the mean contact forces of the candidate and ground-truth executions.
In Tab.~\ref{tab:main_interhuman}, the first seven metrics are computed directly on the generated digital interactions, whereas Force Error is evaluated at the simulator level under the same pretrained tracker. Tab.~\ref{tab:interagent} reports all metrics computed on simulated humanoid trajectories.
To compute the text-motion consistency metrics, we separately train two motion-text encoder pairs with the contrastive loss~\citep{petrovich2023tmr}: one on text-motion pairs and the other on text-humanoid pairs. Full definitions and more details are provided in Appendix~\ref{app:metric} and Appendix~\ref{app:tmr}.

\textbf{Implementation Details.}
Following InterGen~\citep{liang2024intergen} and TIMotion~\citep{wang2025timotion}, we adopt their human--human interaction models as the diffusion backbones. For each training prompt in InterHuman and Inter-X, we sample $K=12$ candidate motions from the corresponding diffusion model. We use NVIDIA Isaac Gym~\citep{makoviychuk2021isaac} as the physics simulation environment, in which the candidates are evaluated through tracking rollouts to construct preference pairs.  During training, \ours~is optimized using the AdamW optimizer with a batch size of 32, and the learning rate is set to $2\times10^{-5}$. During inference, we generate motions over 50 steps using the DDIM strategy and classifier-free text guidance.

\begin{figure}[t]
    \centering
    \includegraphics[width=0.92\linewidth]{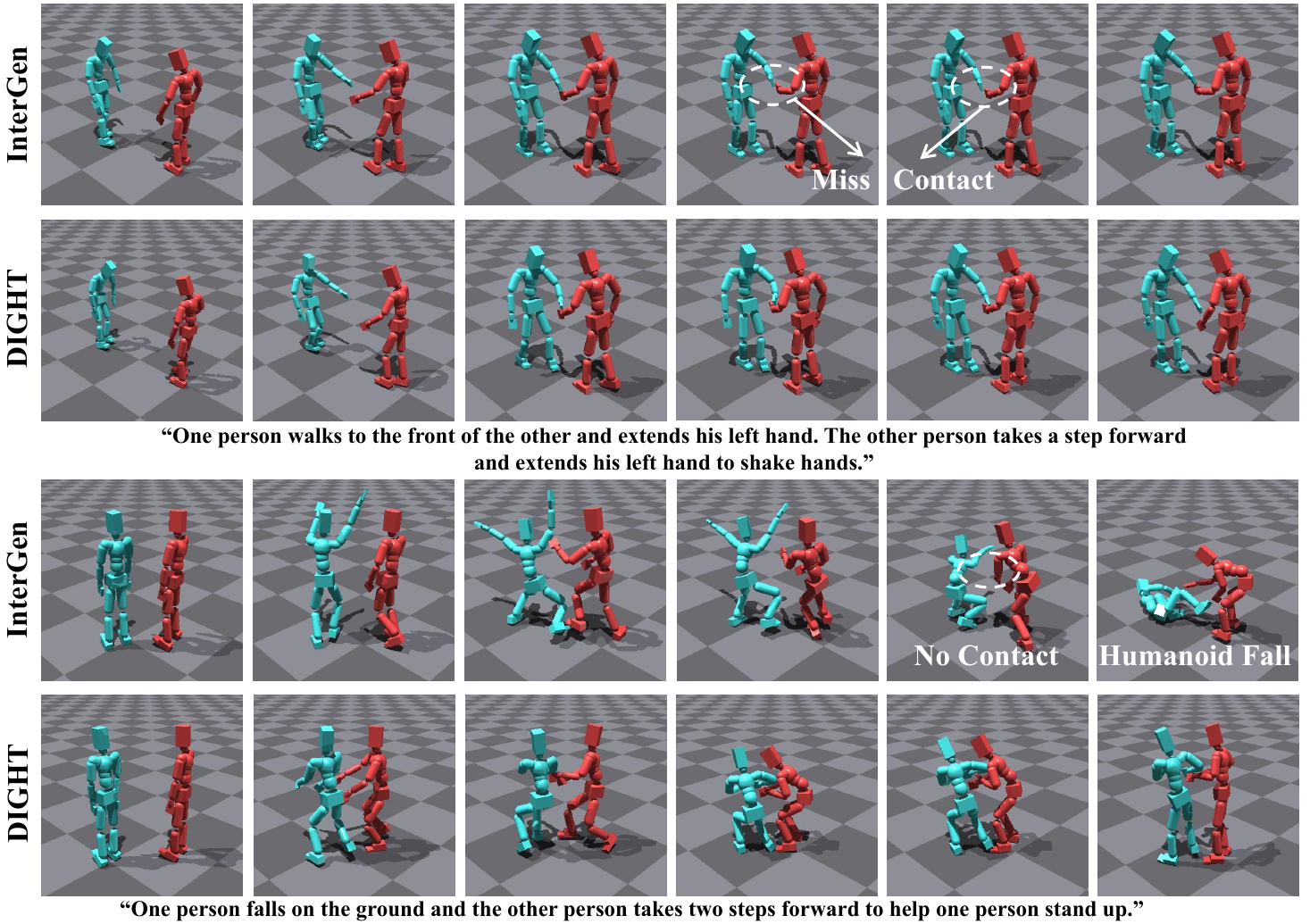}
    \vspace{-4mm}
    \caption{\textbf{Qualitative comparison.} We compare our \ours~with InterGen under various text prompts. Our method generates more natural and physically plausible humanoid interactions, with more realistic and coherent physical contacts, while better aligning with the given text.}
    \vspace{-5mm}
    \label{fig:comp}
\end{figure}

\subsection{Generator Evaluation}
\textbf{Quantitative Results.}
Tab.~\ref{tab:main_interhuman} presents the quantitative results of the generator in \ours~on the InterHuman. To demonstrate the generality of our framework, we instantiate \ours~with two different motion generators, InterGen and TIMotion, denoted as {\ours}$^1$ and {\ours}$^2$, respectively. Across both backbones, our method consistently reduces all physical artifacts while largely preserving motion quality and text-motion alignment. {\ours}$^1$ and {\ours}$^2$ reduces the five physical metrics by averages of 22.1\% and 25.3\%, respectively. These results demonstrate that physics-grounded preferences introduce physical supervision into the digital generator, improving the physical plausibility of its generated results. The quantitative results of Inter-X are shown in Appendix~\ref{app:inter_x}.

\textbf{Qualitative Results.}
In Fig.~\ref{fig:comp}, we present visual comparisons between \ours~and the InterGen baseline. It can be seen that InterGen struggles to produce physically plausible interactions when executed in the simulation environment, often failing to establish intended contacts (e.g., during handshaking) or offer necessary physical force to support (e.g., when helping a falling person stand up). In contrast, our approach is able to generate motions that better preserve the intended contacts, while maintaining more natural and physically plausible interactions between the two humanoid. These observations demonstrate the superior ability of \ours~to generate physically plausible human-human interactions. Please find more visual comparison results in the Appendix~\ref{app:vis}.

\begin{table*}[t]
\centering

\vspace{0pt}
\begin{minipage}[t]{0.53\linewidth}
\centering
\caption{\textbf{Ablation study of different preference combinations on InterHuman test set.}}
\label{tab:reward}
\begin{adjustbox}{width=\linewidth}
\footnotesize
\begin{tabular}{l|ccccc}
\toprule
Reward & Skate & Float & Interpen & Jerk & Force Error \\
\midrule
- & 7.27 & 22.40 & 10.69 & 148.96 & 577.41 \\
trk & 6.74 & 20.12 & 9.93 & 138.16 & 528.37 \\
+ fric & 5.71 & 19.85 & 9.44 & 130.21 & 511.82 \\
+ float & 5.93 & 18.81 & 9.51 & 124.57 & 496.49 \\
+ contact & \textbf{5.67} & \textbf{18.58} & \textbf{8.42} & \textbf{111.94} & \textbf{427.98} \\
\bottomrule
\end{tabular}
\end{adjustbox}
\end{minipage}
\hfill
\begin{minipage}[t]{0.43\linewidth}
\centering
\vspace{0pt}
\caption{\textbf{Performance in the simulation environment.} $\text{\ours}^{*}$ denotes the result without fine-tuning the tracking policy.}
\label{tab:sim}
\label{tab:simulation}
\begin{adjustbox}{width=\linewidth}
\footnotesize
\begin{tabular}{l|ccc}
\toprule
Method & Succ. (\%) & Fall (\%) & MBD (cm) \\
\midrule
InterGen & 88.73 & 9.32 & 10.36 \\
$\text{\ours}^{*}$ & 91.45 & 6.28 & 8.22 \\
$\text{\ours}$ & \textbf{92.19} & \textbf{5.77} & \textbf{7.86} \\
\bottomrule
\end{tabular}
\end{adjustbox}
\end{minipage}
\vspace{-5mm}
\end{table*}

\subsection{Humanoid Execution Evaluation}
\textbf{Comparison with Direct Humanoid Generation.}
We further compare \ours~with InterAgent~\citep{li2026interagent}, a state-of-the-art method for directly generating humanoid motions. As shown in Tab.~\ref{tab:interagent}, \ours~consistently outperforms InterAgent across all evaluation metrics, achieving better motion quality and text alignment while producing more physically plausible interactions.

\begin{wraptable}{r}{0.45\linewidth}
  \centering
  \vspace{-23pt}
  \caption{\textbf{Performance on text prompts requiring physical contact.}}
  \label{tab:contact}
  \begin{adjustbox}{width=\linewidth}
    \footnotesize
    \begin{tabular}{l|ccc}
      \toprule
      Method & Topo. Prec. & Topo. Recall & Succ. \\
      \midrule
      InterGen & 38.29 & 52.16 & 84.70 \\
      \ours & \textbf{46.77} & \textbf{63.94} & \textbf{89.23} \\
      \bottomrule
    \end{tabular}
  \end{adjustbox}
  \vspace{-10pt}
\end{wraptable}
\textbf{Contact Fidelity.}
We further evaluate \ours~on text prompts requiring physical contact. To obtain the annotations required for this evaluation, we use GPT-5.6-sol to extend the InterHuman test set by determining from each textual description whether physical contact is required and which body-region pairs are expected to make contact.
Detailed annotation procedures are provided in the Appendix~\ref{app:gpt_detail}. 
As shown in Tab.~\ref{tab:contact}, compared to baseline, \ours~increases contact topology precision from 38.29\% to 46.77\% and recall from 52.16\% to 63.94\%. Meanwhile, the success tracking rate of contact motion increases from 84.70\% to 89.23\%. These improvements indicate that the proposed contact-related preference optimization enables more accurate and reliable contact interactions, rather than simply increasing unnecessary contacts.

\textbf{Effect of Generator-tracker Co-adaptation.}
Tab.~\ref{tab:sim} evaluates the execution performance of the baseline, \ours~without tracker adaptation, and the complete \ours. Aligning the generator through physics-grounded preference optimization raises execution success from 88.73\% to 91.45\%, reduces fall rate from 9.32\% to 6.28\%, and lowers tracking error (MBD) from 10.36 cm to 8.22 cm. Adapting the tracker to this updated generator provides further gains on all three metrics.
These results show that physics-grounded preference optimization produces more physically plausible interactions that are easier for the humanoid tracker to execute, while adapting the tracker to the updated motion distribution further improves execution robustness and tracking accuracy. 
Notably, our framework naturally supports iterative generator-tracker co-adaptation: the adapted tracker can be used to collect updated physical feedback for the next round of generator optimization, followed by further tracker adaptation. This iterative strategy is optional. We use a single round co-adaptation in the main experiments and study the effect of iterative strategy in the Appendix~\ref{app:iterative}.

\subsection{Discussion}
\textbf{Effect of Different Physics-grounded Preferences.}
To study the contribution of each preference criterion, we progressively incorporate tracking, friction, floating, and contact preferences into generator optimization. As shown in Tab.~\ref{tab:reward}, the tracking preference improves all five physical metrics, while adding friction and floating preferences further reduces the corresponding skating and floating artifacts. Finally, adding the contact reward yields the best overall performance. These results demonstrate that the different physics-grounded preferences provide complementary supervision for improving both general executability and interaction fidelity.

\textbf{Effect of Physics-decoupled DPO Strategy.}
To evaluate the effectiveness of our proposed physics-decoupled DPO strategy, we compare it against standard DPO and supervised fine-tuning (SFT). As shown in Tab~\ref{tab:main_interhuman}, SFT leads to degradation in semantic alignment, motion quality and the physical metric of interpenetration. Standard DPO improves physical plausibility to some extent, but the gains remain relatively modest. In contrast, our method consistently outperforms both the SFT and the standard DPO in all evaluation metrics, demonstrating the effectiveness of the proposed physics-decoupled optimization strategy.

\begin{figure}
    \centering
    \includegraphics[width=0.9\linewidth]{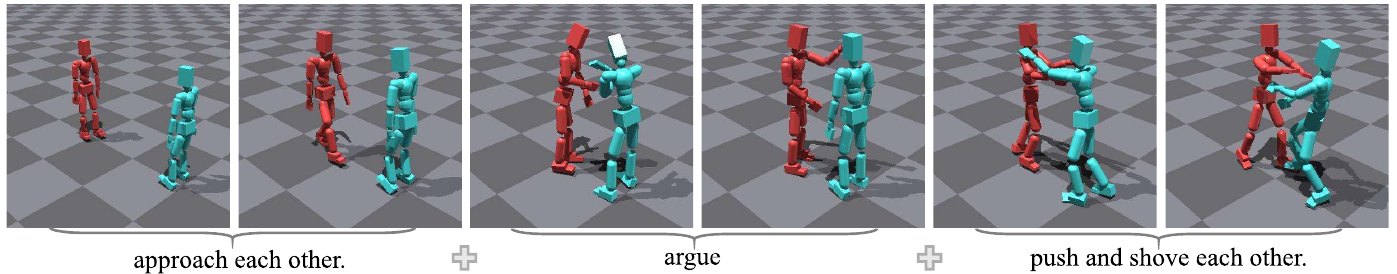}
    \vspace{-4.5mm}
    \caption{Qualitative results of long-horizon humanoid interaction generation}
    \vspace{-4mm}
    \label{fig:open_world}
\end{figure}
\begin{figure}
    \centering
    \includegraphics[width=0.9\linewidth]{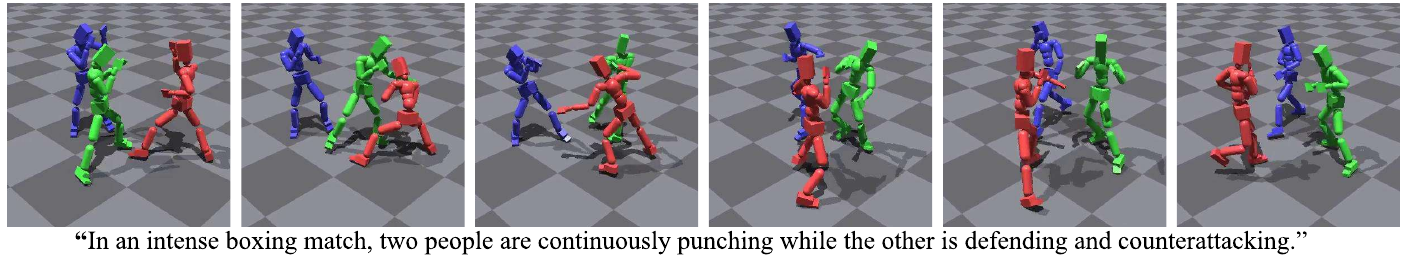}
    \vspace{-4.5mm}
    \caption{Qualitative results of multi-humanoid interaction generation.}
    \vspace{-6mm}
    \label{fig:multiperson}
\end{figure}

\textbf{Application Extensions.}
We further evaluate \ours~ on two challenging applications: long-horizon temporal composition and multi-humanoid interaction. For long-horizon interaction, following PriorMDM~\citep{shafir2024human}, we apply temporal composition to the generator to synthesize a continuous two-person reference motion from multiple temporally ordered instructions, which is subsequently executed by the adapted humanoid tracker. As shown in Fig.~\ref{fig:open_world}, two people walk toward each other, meet face to face, argue, and eventually push one another. The physical execution preserves this temporal progression and produces a coherent long-horizon interaction. We also investigate an $N=3$ scenario beyond the pairwise setting. Following Xu et al.~\citep{xu2025multi}, given the prompt ``Three humanoids boxing'', \ours~produces three humanoids simultaneously participate in the interaction. As shown in Fig.~\ref{fig:multiperson}, the humanoids change their relative configurations and perform distinct boxing-like motions toward different partners. These results demonstrate the extensibility of \ours~to both temporally composed and multi-humanoid interactions.

\section{Conclusion}
In this paper, we propose \ours, a two-stage co-adaptive framework consisting of an interaction generator and a humanoid tracking policy. Using simulator feedback, \ours~ constructs physics-grounded preferences for general executability and interaction fidelity, with the latter explicitly modeling the contact occurrence, duration, and interaction forces. These preferences post-train the generator through physics-decoupled diffusion DPO. The aligned generator is subsequently used to adapt the tracking policy to the updated motion distribution. Extensive experiments demonstrate that \ours~improves physical executability and interaction fidelity.

\clearpage

\subsection*{AI use statement}
In this work, we used OpenAI's GPT model for two purposes. First, GPT-5.6-Sol was used to generate initial annotations for a dataset. All AI-generated annotations were subsequently reviewed item by item by human annotators and corrected where necessary. Second, we use the ChatGPT to polish the language of the manuscript, including its grammar, clarity, and readability. It was not used to generate or alter the research ideas, hypotheses, methodology, experimental design, results, interpretations, scientific claims, or conclusions of this work. Other tasks listed as requiring disclosure under the ICLR policy were either not performed with generative AI or were not applicable to this work. We reviewed all AI-assisted work and take full responsibility for the final content of the paper and the accompanying dataset.

\bibliography{iclr2027_conference}
\bibliographystyle{iclr2027_conference}

\newpage
\appendix
\section*{Appendix}
In this appendix, we present the following:
\begin{itemize}
    \item Section~\ref{app:vis}: More visualization results \\
    \item Section~\ref{app:inter_x}: Results on Inter-X\\
    \item Section~\ref{app:iterative}: Iterative optimization strategy. \\
    \item Section~\ref{app:method-details}: Additional method details\\
    \item Section~\ref{app:metric}: Full metrics definition.\\
    \item Section~\ref{app:dataset-details}: More details of datasets.\\
    \item Section~\ref{app:gpt_detail}: Annotated contact on InterHuman testset. \\
    \item Section~\ref{app:motion-retarget}: Motion retarget.\\
    \item Section~\ref{app:tracking}: Tracking policy.\\    
    \item Section~\ref{app:tmr}: Text-motion evaluators.\\
\end{itemize}

\section{More Visualization Results}
\label{app:vis}
In this section, we provide additional visual comparisons, a long-horizon humanoid interaction sequence, multi-humanoid interaction results, a highly dynamic interaction result, and a rich contact interaction result. Please refer to \href{https://1233chen.github.io/DIGHT/}{\textit{project page}} for video results and additional comparisons with InterGen.

\begin{figure}[t]
    \centering
    \includegraphics[width=0.95\linewidth]{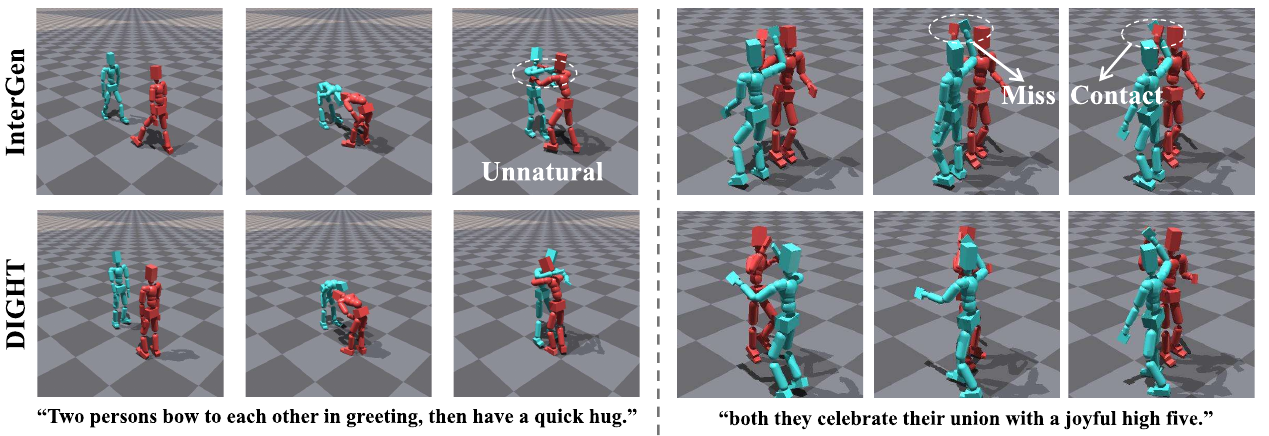}
    \caption{Additional qualitative comparisons between our method and the baseline.}
    \label{fig:app_viscomp1}
\end{figure}

\begin{figure}[t]
    \centering
    \includegraphics[width=0.95\linewidth]{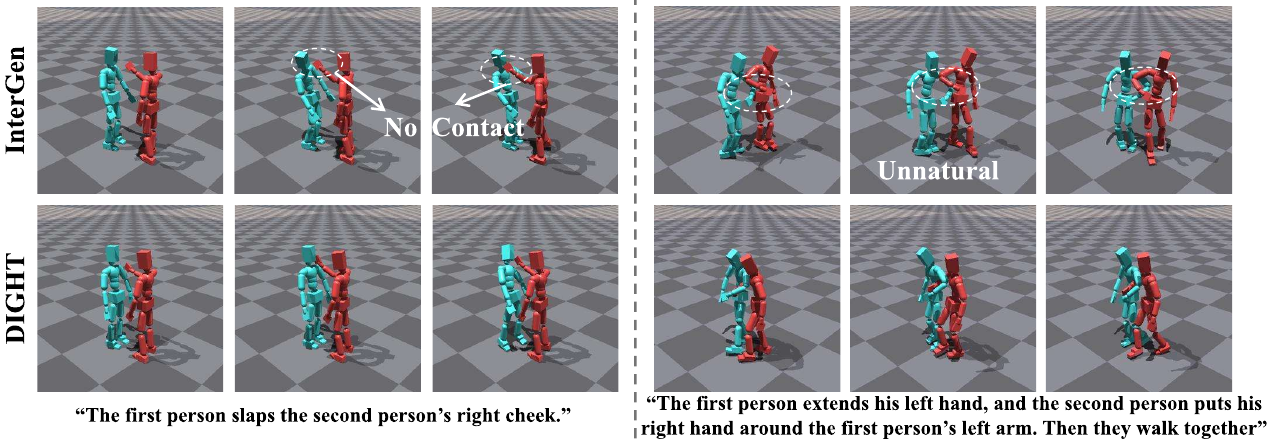}
    \caption{Additional qualitative comparisons between our method and the baseline.}
    \label{fig:app_viscomp2}
\end{figure}

\begin{figure}[t]
    \centering
    \includegraphics[width=0.95\linewidth]{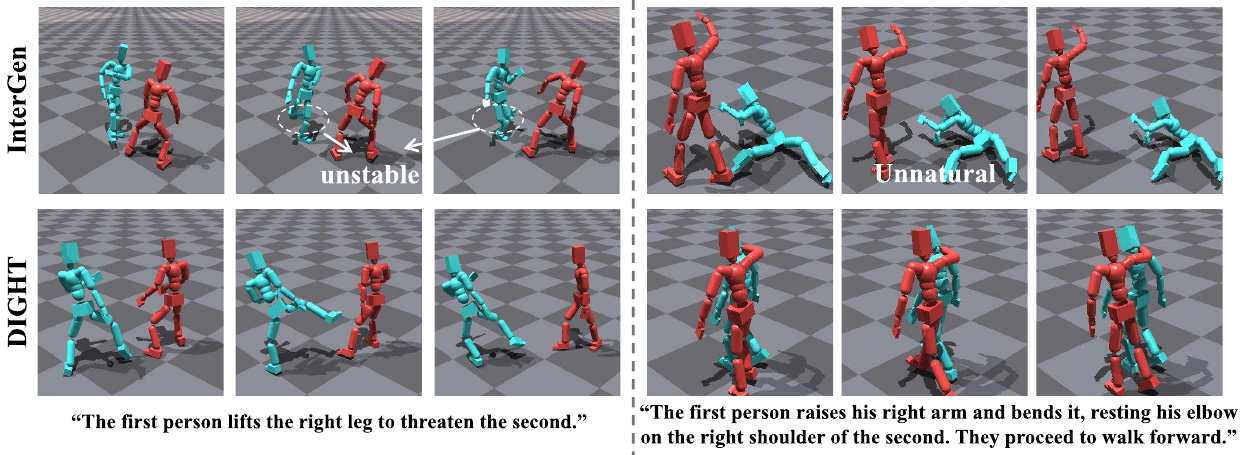}
    \caption{Additional qualitative comparisons between our method and the baseline.}
    \label{fig:app_viscomp3}
\end{figure}

\textbf{More Comparison Results.}
As shown in Figs.~\ref{fig:app_viscomp1}, \ref{fig:app_viscomp2}, and \ref{fig:app_viscomp3}, we provide six additional sets of qualitative comparisons with the baseline. These examples further demonstrate that our method produces more physically plausible motions while better preserving the intended interactions.

\begin{figure}[t]
    \centering
    \includegraphics[width=1.0\linewidth]{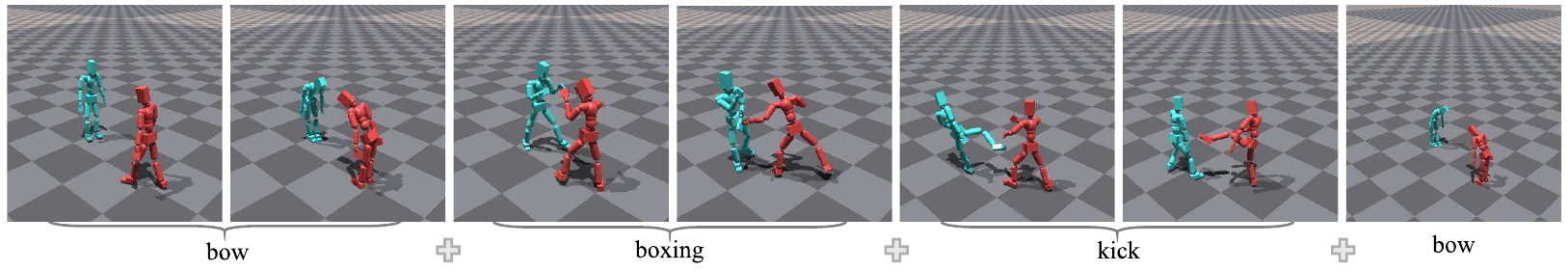}
    \caption{Additional long-horizon humanoid interaction result.}
    \label{fig_app:long_motion}
\end{figure}

\textbf{Long-horizon Humanoid Interaction Result.}
As shown in Fig.~\ref{fig_app:long_motion}, we present a long-horizon interaction sequence composed of multiple dynamically diverse actions. The resulting humanoid motion exhibits coherent transitions between the constituent actions and remains physically stable throughout the sequence.

\begin{figure}[t]
    \centering
    \includegraphics[width=0.95\linewidth]{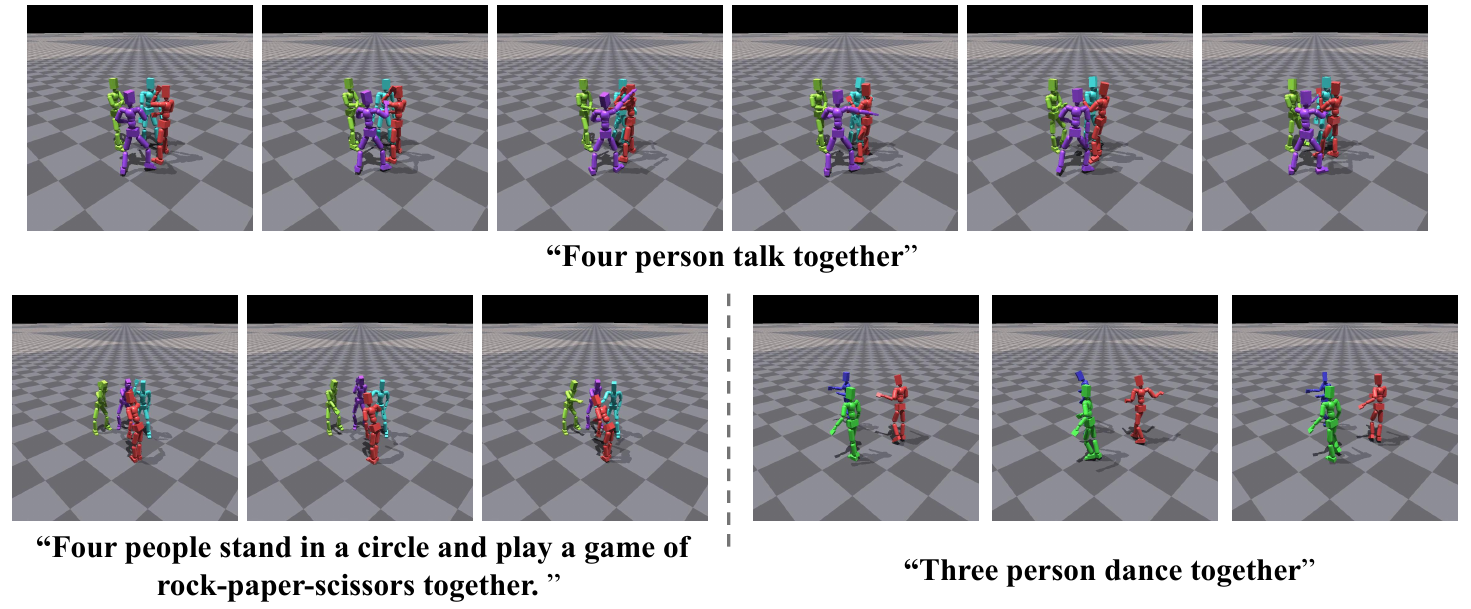}
    \caption{Additional multi-humanoid interaction results.}
    \label{fig_app:multi_person}
\end{figure}

\textbf{Multi-humanoid Interaction Results.}
As shown in Fig.~\ref{fig_app:multi_person}, we additionally present three interaction scenarios involving three or four humanoids. Our method successfully generates and executes these multi-humanoid interactions while maintaining coordinated behavior among all participants.

\begin{figure}[t]
    \centering
    \includegraphics[width=1.0\linewidth]{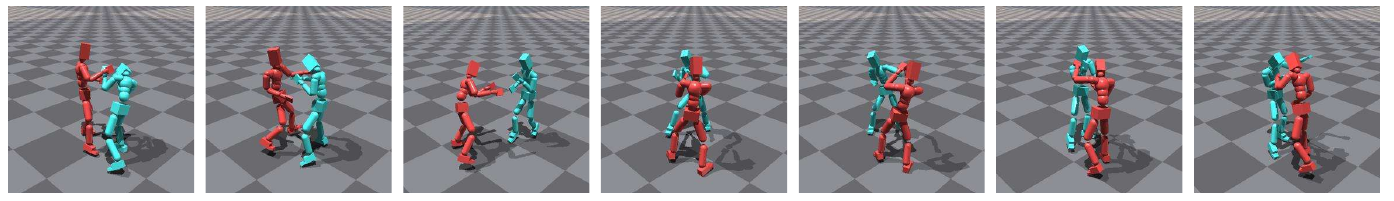}
    \caption{An appealing humanoid interaction with sustained contact (boxing). Please refer to the supplementary video.}
    \label{fig:box}
\end{figure}

\begin{figure}[t]
    \centering
    \includegraphics[width=1.0\linewidth]{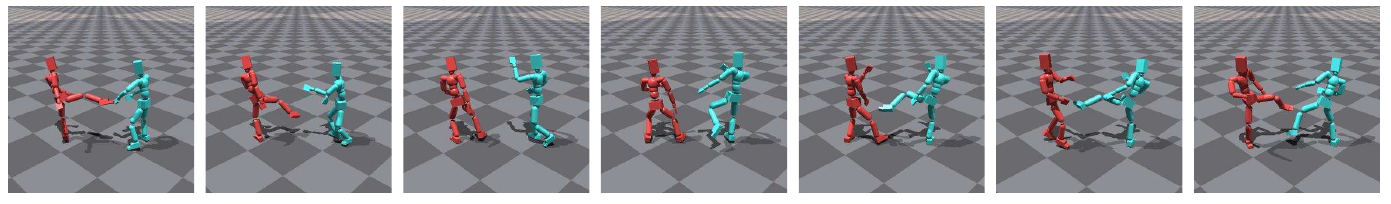}
    \caption{An appealing humanoid interaction result featuring a highly dynamic motion (kick).}
    \label{fig:kick}
\end{figure}

\textbf{Highly Dynamic and Contact-rich Interaction Results.}
As shown in Figs.~\ref{fig:box} and \ref{fig:kick}, we additionally present a boxing sequence with sustained contact and a highly dynamic kicking sequence. The boxing example demonstrates coordinated motion during prolonged close-range contact, while the kicking example involves rapid limb movements, substantial shifts in balance, and dynamic contact. In both cases, our method produces physically stable motions while preserving the intended interactions.

\section{Results on Inter-X}
\label{app:inter_x}
To further demonstrate the generalizability of our framework, we conduct experiments on the Inter-X~\citep{xu2024inter} dataset using both InterGen~\citep{liang2024intergen} and TIMotion~\citep{wang2025timotion} as generator backbones. As shown in Tab.~\ref{tab:main_interx}, \ours~consistently improves all five physical metrics across both backbones while largely preserving motion quality and text-motion alignment.  When initialized from InterGen, \ours~reduces the five physical errors by an average of 16.6\%. Similar improvements are observed with TIMotion, where the average relative reduction reaches 18.3\%. These consistent gains across a different interaction dataset and two distinct generator architectures demonstrate that our physics-grounded alignment framework generalizes beyond a specific dataset or backbone.

\begin{table*}[t!]
\centering
\caption{\textbf{Quantitative comparisons on Inter-X test set.} The best and second best results are \textbf{bold} and \underline{underlined}. $\text{\ours}^{1}$ and $\text{\ours}^{2}$ denote our framework initialized from InterGen and TIMotion, respectively. The Force Error is evaluated from simulated executions using the same pretrained tracker, while the other metrics evaluate the outputs of the generators.}
\label{tab:main_interx}
\begin{adjustbox}{width=1.0\linewidth}
\begin{tabular}{l|ccc|ccccc}
\toprule
\multirow{2}{*}{Methods} & \multirow{2}{*}{FID\ $\downarrow$}& \multirow{2}{*}{MM Dist\ $\downarrow$} & {R-Precision} & \multirow{2}{*}{Skate\ $\downarrow$} & \multirow{2}{*}{Float $\downarrow$} & \multirow{2}{*}{Interpen $\downarrow$} & \multirow{2}{*}{Jerk $\downarrow$} & \multirow{2}{*}{Force Error $\downarrow$}\\
&&& (Top-1) \ $\uparrow$&&&&& \\
\midrule
InterGen& 1.162 & 0.884 & 0.578 & 4.42 & 14.43 & 16.01 & 63.04 & 236.54 \\
InterGen + SFT & 1.639 & 0.897 & 0.570 & 3.89 & 14.35 & 21.60 & 59.47 & 347.07 \\
InterGen + DPO & 1.125 & 0.886 & 0.578 & 4.17 & 12.66 & 14.97 & 57.33 & 223.68 \\
$\text{\ours}^{1}$& 1.107 & 0.883 & 0.581 & \textbf{3.62} & \textbf{10.87} & \textbf{13.94} & \textbf{52.88} & \textbf{209.92} \\
\midrule
TIMotion & 1.054 & 0.863 & 0.599 & 7.65 & 15.39 & 13.77 & 97.52 & 268.13 \\
$\text{\ours}^{2}$& 1.013 & 0.865 & 0.594 & 5.97 & 12.42 & 11.59 & 76.80 & 232.86 \\
\bottomrule
\end{tabular}
\end{adjustbox}
\end{table*}

\section{Iterative optimization strategy.}
\label{app:iterative}
\begin{wraptable}{r}{0.46\linewidth}
  \vspace{-10pt}
  \centering
  \caption{\textbf{Effect of iterative optimization.}}
  \label{tab:iter}
  \setlength{\tabcolsep}{4pt}
  \renewcommand{\arraystretch}{1.05}
  \footnotesize
  \begin{tabular}{l|ccc}
    \toprule
    Iter. & Succ. (\%) & Fall (\%) & MBD (cm) \\
    \midrule
    1 & 92.19 & 5.77 & 7.86 \\
    2 & \underline{92.88} & \underline{5.31} & \textbf{7.32} \\
    3 & \textbf{93.05} & \textbf{4.96} & \underline{7.41} \\
    4 & 92.51 & 5.48 & 7.68 \\
    \bottomrule
  \end{tabular}
  \vspace{-8pt}
\end{wraptable}
In main paper, we update the generator from the physical feedback of tracker and then adapts the tracker once, but the same two-stage procedure can optionally be repeated. As shown in Tab.~\ref{tab:iter}, increasing the iterations from 1 to 3 improves the success rate from 92.19\% to 93.05\% and reduces the fall rate from 5.77\% to 4.96\%, while MBD decreases from 7.86 cm to 7.41 cm. Further increasing the iterations to 4 slightly degrades the performance. Given that the single iteration already achieves strong results, while additional iterations provide comparatively modest gains and require renewed simulation and optimization, we use the single iteration in our main experiments and treat iteration refinement as an optional extension.

\section{Additional Method Details}
\label{app:method-details}
\subsection{General Physical Preferences}
\label{app:physical-preferences}
For each execution trajectory $\tau$. Let $q_{t}^{p,b}$ and $\bar q_{t}^{p,b}$ denote the executed and reference positions of body $b\in\mathcal{B}$ for participant $p\in\{A,B\}$ at frame $t$. The tracking error is the mean body deviation
\begin{equation}
  E_{\mathrm{trk}}(\tau)
  =
  \frac{1}{2T|\mathcal{B}|}
  \sum_{p\in\{A,B\}}
  \sum_{t=1}^{T}
  \sum_{b\in\mathcal{B}}
  \left\|q_{t}^{p,b}-\bar q_{t}^{p,b}\right\|_2.
\end{equation}
Let $D_{t}^{p,b}$ be the tangential contact dissipation reported by the simulator for an active ground-contact body $b\in\mathcal{F}$. We aggregate this physical loss as
\begin{equation}
  E_{\mathrm{fric}}(\tau)
  =
  \frac{1}{2T}
  \sum_{p\in\{A,B\}}
  \sum_{t=1}^{T}
  \sum_{b\in\mathcal{F}}D_{t}^{p,b}.
\end{equation}
Let $z_{t}^{p,b}$ be the body height in the ground coordinate at frame $t$, the Float metric is
\begin{equation}
    E_{\mathrm{float}}
    =
    \frac{1}{2T}
    \sum_{p,u}
    \max\left(
    \min_{j\in\mathcal{J}}
    z^{p}_{u,j}
    -
    z_{\mathrm{floor}},
    0
    \right).
\end{equation}
We also measure text-motion distance $d_(x,c)$ between the generated motion $x$ and corresponding text $c$ using the frozen text feature and motion feature encoders $f_{\mathrm{motion}}$ and $f_{\mathrm{text}}$,
\begin{equation}
  d_(x,c)
  =
  \left\|f_{\mathrm{motion}}(x)-f_{\mathrm{text}}(c)\right\|_2.
\end{equation}
For target criterion $k$, a pair $(i,j)$ is valid only if
$E_k(\tau_i)<E_k(\tau_j)$ and
$d(x_i,c)<d_(x_j,c)$.

\subsection{Contact Representation and Comparison}
\label{app:contact-preferences}
As defined in the main paper, we use $C(\tau)$ denoting the persistent region pair set for trajectory $\tau$. To calculate the contact quality, we define the normalized topology discrepancy between region pair sets $A$ and $B$ as:
\begin{equation}
  d_{\mathrm{topo}}(A,B)
  =1-
  \frac{|A\cap B|}{|A\cup B|},
  \label{eq:topology_discrepancy}
\end{equation}
with value zero when both sets are empty. And define the bounded log error as:
\begin{equation}
  d_{\log}(a,b;\epsilon)
  =
  \frac{
  |\log(a+\epsilon)-\log(b+\epsilon)|
  }{
  |\log(a+\epsilon)-\log(b+\epsilon)|+\log 2
  }.
  \label{eq:normalized_log_error}
\end{equation}

\noindent\textbf{Contact-topology Error.}
For each humanoid trajectory $\tau$, its contact-topology error is therefore:
\begin{equation}
  E_{\mathrm{topo}}(\tau)
  =
  d_{\mathrm{topo}}\!\left(
  \mathcal{C}(\tau),
  \mathcal{C}(\tau^{\mathrm{gt}})
  \right),
\end{equation}
where $\mathcal{C}(\tau)$ as defined in the main paper denotes the persistent region pairs set.

\noindent\textbf{Edge-wise Contact Duration and Force.}
Let $\mathcal{R}=\{\text{core},\text{head},\text{arm}, \text{hand},\text{leg},\text{foot}\}$ denote the six body regions. Each region pair $e=(r^A,r^B)\in\mathcal{R}\times\mathcal{R}$ corresponds to one entry of the $6\times6$ contact graph.
For each edge $e$, we group its active frames into contact episodes. Before extracting the episodes, we fill inactive gaps of fewer than four frames; thus, two contact segments separated by one to three inactive frames are merged into the same episode. We denote the longest resulting episode in rollout $\tau$ by $I_e^*(\tau)$. Its contact duration is
\begin{equation}
    D_e(\tau)
    =
    |I_e^*(\tau)|.
    \label{eq:edge_contact_duration}
\end{equation}
We set $D_e(\tau)=0$ if edge $e$ has no contact episode.

Let $\mathcal{P}_{e,t}$ denote the active rigid-body pairs mapped to region edge $e$ at frame $t$. The aggregate force of this region edge is
\begin{equation}
    F_{e,t}(\tau)
    =
    \sum_{(i,j)\in\mathcal{P}_{e,t}}
    \left\|
      \sum_{m\in\mathcal{M}_{i,j,t}}
      \mathbf f_{m,t}
    \right\|_2,
    \label{eq:edge_contact_force}
\end{equation}
where $\mathcal{M}_{i,j,t}$ contains the contact manifolds between rigid bodies $i$ and $j$. Each contact force is included only once.
We characterize the force of edge $e$ by its mean and maximum values within its longest contact episode:
\begin{equation}
    \bar F_e(\tau)
    =
    \frac{1}{|I_e^*(\tau)|}
    \sum_{t\in I_e^*(\tau)}
    F_{e,t}(\tau),
    \qquad
    F_e^{\max}(\tau)
    =
    \max_{t\in I_e^*(\tau)}
    F_{e,t}(\tau).
    \label{eq:edge_force_statistics}
\end{equation}
Both statistics are set to zero if $I_e^*(\tau)$ is empty.

Let $\mathcal{C}^{\mathrm{gt}} =\mathcal{C}(\tau^{\mathrm{gt}})$ denote the persistent contact edges of the reference rollout. The edge-wise duration error is
\begin{equation}
    E_{\mathrm{dur}}(\tau)
    =
    \frac{1}{|\mathcal{C}^{\mathrm{gt}}|}
    \sum_{e\in\mathcal{C}^{\mathrm{gt}}}
    d_{\log}
    \left(
      D_e(\tau),
      D_e(\tau^{\mathrm{gt}});
      \epsilon_d
    \right).
    \label{eq:edge_duration_error}
\end{equation}

The edge-wise force error is
\begin{equation}
    E_{\mathrm{force}}(\tau)
    =
    \frac{1}{2|\mathcal{C}^{\mathrm{gt}}|}
    \sum_{e\in\mathcal{C}^{\mathrm{gt}}}
    \left[
      d_{\log}
      \left(
        \bar F_e(\tau),
        \bar F_e(\tau^{\mathrm{gt}});
        \epsilon_f
      \right)
      +
      d_{\log}
      \left(
        F_e^{\max}(\tau),
        F_e^{\max}(\tau^{\mathrm{gt}});
        \epsilon_f
      \right)
    \right].
    \label{eq:edge_force_error}
\end{equation}
Extra contact edges generated by the candidate are penalized separately by the topology error.
Candidates that fall during simulation are excluded from all preference sets except the $\text{trk}$ preference set

\subsection{Diffusion Parameterization}
\label{app:diffusion-objective}
Given a clean two-person interaction $x_0=(x_0^A,x_0^B)$, the forward process samples
\begin{equation}
  q(x_t\mid x_0)
  =
  \mathcal{N}\!\left(
  \sqrt{\bar\alpha_t}\,x_0,
  (1-\bar\alpha_t)I
  \right).
\end{equation}
Both generator backbones predict the clean interaction,
$\hat{x}_0=G_\theta(x_t^A,x_t^B,t,c)$, and iterative reverse diffusion defines $p_\theta(x_0\mid c)$. For a DPO pair, the preferred and rejected motions share the sampled timestep $t$ and are independently perturbed as
\begin{equation}
  x_t^{\pm}
  =
  \sqrt{\bar\alpha_t}\,x^{\pm}
  +
  \sqrt{1-\bar\alpha_t}\,\varepsilon^{\pm},
  \qquad
  \varepsilon^{\pm}\sim\mathcal{N}(0,I).
\end{equation}

\section{Full Metrics Definitions}
\label{app:metric}
\subsection{Motion Quality.}
We follow the evaluation protocols established in prior work~\citep{liang2024intergen, li2026interagent}, and train evaluators as described in Appendix~\ref{app:tmr} to assess the alignment between the generated stylized motions and the input content text.
Specifically, we use Frechet Inception Distance (FID) to assess the visual quality of generated motions, Retrieval Precision (R-Precision) and Multi-Modal Distance (MM-Dist) to evaluate the alignment between motions and the input text.

\noindent\textbf{FID}. 
Fréchet Inception Distance (FID) is employed to evaluate the quality of generated motions by comparing the distribution of motion features between generated and real motion data.
Lower FID values indicate that the generated motions are closer to real samples in terms of visual realism and temporal coherence. 
We extract motion features using the same motion encoder trained in App.~\ref{app:tmr}.
We obtain a set of motion features $X$ from the generated motion data, and another set of features $X'$ from the test dataset.
The formulation of FID is defined as follows:
\begin{equation}
    \text{FID} = ||\mu_1-\mu_2||^2 - \text{Trace}(\sigma_1+\sigma_2-2(\sigma_1\sigma_2)^{\frac{1}{2}}),
\end{equation}
where $\mu_1$ and $\mu_2$ represent the mean feature vectors of $X$ and $X'$, respectively, while $\sigma_1$ and $\sigma_2$ denote their corresponding covariance matrices.

\vspace{2mm}
\noindent\textbf{MM-Dist}.
Multi-Modal Distance (MM-Dist) quantifies the semantic distance between the text and the generated motion in a multi-modal embedding space.
Lower MM-Dist values imply better alignment between the motion and the content text prompt.
We use the same evaluator to extract the text feature $t_i$ and motion feature $m_i$ for all the text prompts and generated motion results.
The formulation of MM-Dist as follows:
\begin{equation}
    \text{MM~Dist} = \frac{1}{N_m}\sum_{i=1}^{N_m}||t_i-m_i||,
\end{equation}
where $N_m$ is the total number of generated motions.

\vspace{2mm}
\noindent\textbf{R-Precision.}
R-Precision is employed to evaluate the semantic alignment between generated motions and their corresponding textual descriptions.
For each generated motion, we form a candidate pool that includes its paired textual description along with 95 distractor descriptions randomly sampled from the test set.
Both the motion and textual descriptions are embedded into a shared feature space, and their pairwise cosine distances are computed.
We rank the candidate descriptions according to these distances, and a retrieval is counted as correct if the paired description is found within the top-$k$ positions.

\subsection{Physical Quality.}
\vspace{2mm}
\noindent\textbf{Float}.
Float measures the vertical clearance between the executed human motion and the ground plane. A large Float value indicates that the character tends to hover above the ground, while a lower value indicates better ground alignment. We compute this metric for digital generated motions. For each frame, we select the lowest vertical position among the 22 body joints, excluding the two palm joints, and clamp below-ground values to zero.
The Float metric is defined as:
\begin{equation}
      h_{i,t,a}
      =
      \max\left(
          \min_{j\in\mathcal{J}_{22}}
          z_{i,t,a,j},
          0
      \right),
\end{equation}
\begin{equation}
      \text{Float}
      =
      \frac{1000}{N_m}
      \sum_{i=1}^{N_m}
      \frac{1}{2T_i}
      \sum_{a=1}^{2}
      \sum_{t=1}^{T_i}
      h_{i,t,a},
\end{equation}
where $z_{i,t,a,j}$ denotes the vertical position, in meters, of body joint $j$ of person $a$ at frame $t$ in motion $i$. $\mathcal{J}_{22}$ denotes the set of 22 body joints excluding the two palm joints, $T_i$ denotes the valid length of motion $i$, and $N_m$ is the total number of evaluated motions. The factor of $1000$ converts the result from meters to millimeters. Below-ground positions do not produce negative Float values and are evaluated separately by the interpenetration metric. Lower Float values indicate better physical plausibility.

\vspace{2mm}
\noindent\textbf{Skate.}
Skate measures the horizontal displacement of the feet while they are close to the ground. A physically plausible motion should exhibit little horizontal foot movement during ground contact, whereas large values indicate undesirable foot sliding. Let $\mathcal{F}$ denote the four foot bodies, including the left and right ankles and toes. Here, $i$ indexes the interaction sequence, $a$ indexes the person in sequence $i$, $t$ indexes the frame, and $T_i$ denotes the number of frames in sequence $i$. A foot is considered close to the ground if its vertical position is below $5$ cm:
\begin{equation}
      g_{i,t,a,f}
      =
      \mathbb{I}
      \left[
          z_{i,t,a,f} < 0.05
      \right],
\end{equation}
where $f\in\mathcal{F}$ indexes the foot body, $z_{i,t,a,f}$ denotes its vertical position in meters, and $\mathbb{I}[\cdot]$ is the indicator function. The horizontal displacement of foot $f$ between two consecutive frames is
\begin{equation}
      d_{i,t,a,f}
      =
      \left\|
          \mathbf{p}^{xy}_{i,t+1,a,f}
          -
          \mathbf{p}^{xy}_{i,t,a,f}
      \right\|_2.
\end{equation}
where $\mathbf{p}^{xy}_{i,t,a,f}\in\mathbb{R}^2$ denotes the horizontal position of foot body $f$ belonging to person $a$ at frame $t$ of sequence $i$. We first compute the average grounded-foot displacement for each person:
\begin{equation}
      s_{i,a}
      =
      \frac{
          \displaystyle
          \sum_{t=1}^{T_i-1}
          \sum_{f\in\mathcal{F}}
          g_{i,t,a,f}\,
          d_{i,t,a,f}
      }{
          \displaystyle
          \max\left(
              1,
              \sum_{t=1}^{T_i-1}
              \sum_{f\in\mathcal{F}}
              g_{i,t,a,f}
          \right)
      }.
\end{equation}
where $s_{i,a}$ denotes the average horizontal displacement of the grounded foot bodies of person $a$ in sequence $i$. The maximum operation in the denominator prevents division by zero when no foot body is detected near the ground. The metric is reported in millimeters; therefore, $s_{i,a}$ is multiplied by $1000$ when the foot positions are represented in meters. No division by the frame interval is applied. Lower Skate values indicate less foot sliding and better contact consistency.
  
\vspace{2mm}
\noindent\textbf{Interpen.}
 Interpenetration measures the geometric overlap between the two interaction.  Each body joint is approximated by a sphere with radius $r_b=0.06$ m. For every frame, we compute all pairwise distances between the body joints of the two people:
\begin{equation}
      D_{i,t,j,k}
      =
      \left\|
          \mathbf{p}^{\mathrm{gen}}_{i,t,1,j}
          -
          \mathbf{p}^{\mathrm{gen}}_{i,t,2,k}
      \right\|_2,
\end{equation}
where $\mathbf{p}^{\mathrm{gen}}_{i,t,a,j}\in\mathbb{R}^{3}$ denotes the generated position of joint $j$ of person $a$ at frame $t$ in motion $i$. The minimum cross-person joint distance at frame $t$ is:
\begin{equation}
      d_{i,t}^{\min}
      =
      \min_{
          j\in\mathcal{J}_{24},
          \,k\in\mathcal{J}_{24}
      }
      D_{i,t,j,k},
\end{equation}
where $\mathcal{J}_{24}$ denotes the set of 24 body joints. The frame-level interpenetration proxy is defined as:
\begin{equation}
      I_{i,t}
      =
      \max\left(
          2r_b-d_{i,t}^{\min},
          0
      \right)
      =
      \max\left(
          0.12-d_{i,t}^{\min},
          0
      \right).
\end{equation}
The motion-level Interpenetration is computed by averaging over all valid frames:
\begin{equation}
      I_i
      =
      \frac{1000}{T_i}
      \sum_{t=1}^{T_i}
      I_{i,t},
\end{equation}
where $T_i$ is the valid length of motion $i$, and the factor of $1000$ converts meters to millimeters. The final Interpenetration metric is obtained using a motion-equal average:
\begin{equation}
      \text{Interpen}
      =
      \frac{1}{N_m}
      \sum_{i=1}^{N_m}
      I_i,
\end{equation}
where $N_m$ is the total number of generated motions. Lower Interpenetration values indicate less cross-person body overlap and better kinematic plausibility.
  
\vspace{2mm}
\noindent\textbf{Jerk.}
Jerk measures the temporal smoothness of the motion by quantifying the rate of change of body acceleration. A large Jerk value indicates abrupt changes in body movement and physically implausible high-frequency oscillations, while a lower value indicates smoother and more natural motion. Let $\mathbf{p}_{i,t,a,j}\in\mathbb{R}^{3}$ denote the position, in meters, of body joint $j$ of person $a$ at frame $t$ in motion $i$. The third-order finite difference of the body position is defined as:
\begin{equation}
      \Delta^{3}\mathbf{p}_{i,t,a,j}
      =
      \mathbf{p}_{i,t+3,a,j}
      -
      3\mathbf{p}_{i,t+2,a,j}
      +
      3\mathbf{p}_{i,t+1,a,j}
      -
      \mathbf{p}_{i,t,a,j}.
\end{equation}
Given a frame rate of $f$ Hz, the discrete jerk vector is computed as:
\begin{equation}
      \mathbf{j}_{i,t,a,j}
      =
      f^{3}
      \Delta^{3}\mathbf{p}_{i,t,a,j}.
\end{equation}
We compute the motion-level Jerk by averaging the jerk magnitude over all valid temporal positions, both people, and all 24 rigid bodies:
\begin{equation}
      J_i
      =
      \frac{1}{
          2N_J(T_i-3)
      }
      \sum_{a=1}^{2}
      \sum_{j=1}^{N_J}
      \sum_{t=1}^{T_i-3}
      \left\|
          \mathbf{j}_{i,t,a,j}
      \right\|_2,
\end{equation}
where $N_J=24$ is the number of rigid bodies and $T_i$ is the valid executed length of motion $i$. The final Jerk metric is obtained using a motion-equal average:
\begin{equation}
      \text{Jerk}
      =
      \frac{1}{N_m}
      \sum_{i=1}^{N_m}
      J_i,
\end{equation}
where $N_m$ is the total number of evaluated motions. The metric is reported in $\mathrm{m}/\mathrm{s}^{3}$, and lower Jerk values indicate smoother and more physically plausible motion.
  
\vspace{2mm}
\noindent\textbf{Force Error.}
Force Error evaluates whether the intensity of inter-person physical contacts in the generated motions is consistent with that of the corresponding ground-truth motions. We obtain exact rigid-body contacts from the PhysX simulator after each physics step. For each contact point $c$ associated with an inter-person body pair $e$, PhysX provides a normal contact magnitude $\lambda_c$ and a world-space contact normal $\mathbf{n}_c$. All contact normals are consistently oriented from the first person to the second person. The resultant normal-contact magnitude for body pair $e$ at physics query $q$ is:
\begin{equation}
      r_{i,q,e}
      =
      \left\|
          \sum_{c\in\mathcal{C}_{i,q,e}}
          \lambda_c\mathbf{n}_c
      \right\|_2,
\end{equation}
where $\mathcal{C}_{i,q,e}$ is the set of contact manifold points belonging to body pair $e$. A body pair is considered contact-active when its resultant magnitude exceeds the contact threshold $\tau=1.0$. The total contact force at physics query $q$ is:
\begin{equation}
      L_{i,q}
      =
      \sum_e
      \mathbb{I}
      \left[
          r_{i,q,e}>\tau
      \right]
      r_{i,q,e}.
\end{equation}
Let $\mathcal{Q}^{+}_i=\{q\mid L_{i,q}>0\}$ denote the set of contact-active physics queries. The mean contact force of motion $i$ is:
\begin{equation}
      \mu_i
      =
      \frac{1}{|\mathcal{Q}^{+}_i|}
      \sum_{q\in\mathcal{Q}^{+}_i}
      L_{i,q}.
\end{equation}
We compare the generated and ground-truth mean force to calculate the force error:
\begin{equation}
      e_i^{\mathrm{force}}
      =
          \left|
              \mu_i^{\mathrm{gen}}
              -
              \mu_i^{\mathrm{GT}}
          \right|.
\end{equation}
For multiple generated motions associated with the same text prompt, we first average the errors within each prompt and then average equally over all prompts:
\begin{equation}
      \text{Force Error}
      =
      \frac{1}{N_p}
      \sum_{p=1}^{N_p}
      \frac{1}{|\mathcal{S}_p|}
      \sum_{i\in\mathcal{S}_p}
      e_i^{\mathrm{force}},
\end{equation}
where $\mathcal{S}_p$ denotes the set of valid generated motions associated with prompt $p$, and $N_p$ is the number of evaluated prompts. The final Force Error is measured in Newtons (N), and lower values indicate that the generated contact intensity is closer to that of the tracker-executed ground-truth motion.
  
\vspace{2mm}
\noindent\textbf{Success Rate (Succ.).}
Success Rate (Succ.) measures the fraction of generated two-person motion sequences that can be tracked with sufficiently low average error while remaining stable over the entire sequence. For each generated interaction sequence $i$, we compute the sequence-averaged mean per-joint tracking error of person $p$ as
\begin{equation}
        d_{i,p}
        =
        \frac{1}{T_i}
        \sum_{t=1}^{T_i}
        \frac{1}{J}
        \sum_{j=1}^{J}
        \left\|
        \mathbf{x}_{i,p,t,j}
        -
        \mathbf{x}^{\mathrm{ref}}_{i,p,t,j}
        \right\|_2,
\end{equation}
where $\mathbf{x}_{i,p,t,j}$ and $\mathbf{x}^{\mathrm{ref}}_{i,p,t,j}$ denote the global positions of the $j$-th body joint in the simulated and reference motions, respectively. Here, $p\in\{1,2\}$ denotes the person index, $T_i$ is the number of evaluated frames in the $i$-th motion pair, and $J=24$ is the number of body joints.

A motion pair is considered successful if the sequence-averaged tracking error of each person does not exceed the tracking threshold $\tau_d$ and neither person's pelvis height falls below the stability threshold $\tau_h$ at any frame:
\begin{equation}
      s_i
      =
      \mathbb{I}
      \left[
      \max_{\substack{p\in\{1,2\}}}
      d_{i,p}
      \leq \tau_d
      \;\land\;
      \min_{\substack{p\in\{1,2\}\\1\leq t\leq T_i}}
      z^{\mathrm{pelvis}}_{i,p,t}
      \geq \tau_h
      \right],
\end{equation}
where $\mathbb{I}[\cdot]$ is the indicator function.
The Success Rate is then computed as
\begin{equation}
      \mathrm{Succ.}
      =
      \frac{1}{N_m}
      \sum_{i=1}^{N_m}s_i,
\end{equation}
where $N_m$ is the total number of evaluated two-person motion pairs. Higher Succ. values indicate better full-sequence tracking performance. In our evaluation, we set $\tau_d=0.25\,\mathrm{m}$ and $\tau_h=0.3\,\mathrm{m}$.

\vspace{2mm}
\noindent\textbf{MBD.}
Mean Body Deviation (MBD) measures the average global positional discrepancy between the simulated and reference body motions. For each person in a motion pair, we first average the body-joint distances over all joints and all frames of that motion. We then average the resulting motion-level distances over both persons and all evaluated motion pairs.
The formulation of MBD is as follows:
\begin{equation}
      \mathrm{MBD}
      =
      \frac{1}{N_m}
      \sum_{i=1}^{N_m}
      \left[
      \frac{1}{2}
      \sum_{p=1}^{2}
      \left(
      \frac{1}{T_i}
      \sum_{t=1}^{T_i}
      \frac{1}{J}
      \sum_{j=1}^{J}
      \left\|
      \mathbf{x}_{i,p,t,j}
      -
      \mathbf{x}^{\mathrm{ref}}_{i,p,t,j}
      \right\|_2
      \right)
      \right],
\end{equation}
where $N_m$ is the total number of evaluated two-person motion pairs, $T_i$ is the motion length, and $J=24$ is the number of body joints. Lower MBD values indicate more accurate motion tracking.

We adopt motion-level averaging, such that each motion pair contributes equally to the final MBD regardless of its sequence length.

\section{More Details of Datasets}
\label{app:dataset-details}
Our experiments are conducted on two datasets: InterHuman~\citep{liang2024intergen} and Inter-X~\citep{xu2024inter}.

\noindent\textbf{InterHuman.}
InterHuman is a standard benchmark for text-conditioned two-person motion generation. It contains 6,022 interaction sequences paired with 16,756 natural-language descriptions, amounting to 6.56 hours of motion. Each sequence represents the coordinated motions of two participants in a shared coordinate frame, enabling the evaluation of both individual motion quality and inter-person coordination.

\noindent\textbf{Inter-X.}
Inter-X is a large-scale dataset for versatile human--human interaction analysis. It contains 11,388 SMPL-X interaction sequences and 34,164 natural-language descriptions, spanning 40 interaction categories and more than 8.1 million motion frames. Compared with InterHuman, Inter-X provides broader interaction coverage and more detailed body and hand motions, offering a complementary benchmark for evaluating generalization across interaction distributions.

For both datasets, we follow the official train, validation, and test splits and the preprocessing protocols of the corresponding interaction-generation backbones. To ensure a consistent embodiment between motion generation and humanoid tracking, we retarget all interaction sequences to the neutral SMPL body with zero shape parameters. The retarget details are shown in Appendix~\ref{app:motion-retarget}.

\section{Annotated Contact on InterHuman Testset}
\label{app:gpt_detail}
InterHuman provides paired motions and natural-language descriptions, but it does not specify which cross-person body contacts are semantically required by each description. To support contact-sensitive evaluation, we construct a structured annotation subset of the official InterHuman test set. The annotations describe the contact topology entailed by a caption which are not frame-level contact labels extracted from the ground-truth motion. This distinction is important because the evaluation asks whether a generated physical interaction realizes the contact semantics expressed in the text, while allowing different valid motion realizations.

We process the captions associated with valid motions in the official test split. Motion arrays are used only to establish test-set membership and validity and are never provided to the annotation model.

\paragraph{Structured Contact Specification.}
Each caption is mapped to a typed contact specification over the 24 rigid bodies of the simulated humanoid. In addition to exact body names, the ontology provides semantic selectors such as \texttt{Hand}, \texttt{Arm}, \texttt{Foot}, \texttt{Leg}, \texttt{Trunk}, and \texttt{Any\_Body}, each of which expands deterministically to a set of exact bodies. A specification contains: (i) \emph{required} contact clauses that must be realized; (ii) \emph{allowed} contacts that are explicitly compatible with the caption but not mandatory; and (iii) \emph{forbidden} contacts explicitly ruled out by the caption. Each required clause may contain multiple \texttt{any\_of} alternatives, allowing several body-pair realizations of the same semantic contact, together with a \texttt{distinct\_count} when the caption requires multiple distinct contacts. This representation avoids treating one arbitrarily selected body pair as the unique realization of an ambiguous instruction.

The annotation contract is deliberately conservative. It requires evidence quoted verbatim from the caption and prohibits inferring contact merely from proximity, synchronized motion, reaching, attempted actions, or object transfer. It also records controlled ambiguity flags, an abstention decision, a confidence score, and whether contacts not mentioned by the caption should be ignored or treated as disallowed. Participant labels are directional, but the complete specification is canonicalized under one globally consistent $A/B$ transpose; individual contact edges cannot be swapped independently.

\paragraph{Multi-pass Annotation and Strict Filtering.}
We annotate each exact caption in two separate text-only passes using GPT-5.6-sol. The first pass proposes a contact specification, whereas the second uses a skeptical prompt designed to reject contacts that are plausible in a typical execution but are not entailed by the text. Both outputs must satisfy the same JSON schema and are additionally checked by a local semantic validator. The validator verifies the selector vocabulary, required/allowed/forbidden consistency, valid multiplicities, confidence range, and exact matching between every evidence span and the input caption. A caption is included in the final subset only when the two passes agree and satisfy our confidence and quality requirements. Selected disagreement or high-risk cases undergo blind third-pass adjudication and are included only when consensus is reached under the same strict criteria.
The final strict subset contains 503 contact-needed captions from 223 motions and specifies 631 required contact slots. It comprises 445 annotations accepted by two-pass consensus and 58 additional annotations resolved by the blind third pass. We further manually reviewed all annotations in the final strict subset to ensure their accuracy. All remaining captions are withheld from strict contact evaluation.

\paragraph{Core Annotation Prompt.}
The complete prompt supplies the body ontology and output schema. Its central instruction is:
\begin{quote}
\small\itshape
Given exactly one interaction caption, identify only cross-person body contacts that are explicitly entailed by the text. Do not infer contact from a typical execution, proximity, synchronized motion, reaching, attempted actions, or object transfer. Represent mandatory contacts as required clauses; use \texttt{any\_of} when the caption permits multiple body-pair realizations and \texttt{distinct\_count} when it explicitly requires multiple distinct contacts. Add allowed or forbidden contacts only when supported by the caption. If the contact cannot be determined safely, abstain. For every non-abstaining decision, quote an exact supporting span from the caption and return the result using the provided schema.
\end{quote}

\paragraph{Evaluation Metrics.}
The annotations remain in the 24-body ontology to preserve left/right and fine-grained body identities. For the evaluation in Tab.~\ref{tab:contact}, Topology Recall is the fraction of required slots matched, while Topology Precision measures the fraction of evaluated predicted region pairs that are required or allowed by the specification. Contact success indicates whether the rollout completes without humanoid falling and the mean body deviation is less than $\tau_d$.

\section{Motion Retarget}
\label{app:motion-retarget}

The source interaction data and the simulated humanoids have different body proportions and kinematic structures. We first use General Motion Retargeting (GMR)~\citep{araujo2025retargeting} to map each participant to the target humanoid. The two participants are retargeted in the same world coordinate frame so that their relative placement and timing are retained. Denote the resulting target configurations by $\widetilde{q}_t^p$, where $p\in\{A,B\}$ and $t\in\{1,\ldots,T\}$. Although GMR provides a strong whole-body initialization, morphology differences can still perturb fine-grained inter-person contacts at the limb end effectors. An intended hand or foot contact in the source motion may become separated after retargeting or displaced to a nearby joint on the target humanoid.

\paragraph{Source End-effector Contacts.}
We infer the intended end-effector contacts before retargeting, rather than treating every pair of nearby joints as a contact constraint. Let $j_{t,i}^{p,\mathrm{src}}\in\mathbb{R}^3$ denote the position of joint $i$ for participant $p\in\{A,B\}$ in the source motion. We define the limb end-effector set $\mathcal{J}_{\mathrm{ee}}$, containing the left and right hand and foot joints, together with a selected set of potential counterpart joints $\mathcal{J}_{\mathrm{c}}$. Candidate pairs are restricted to
\begin{equation}
 \mathcal{P}_{\mathrm{ee}}
 =
 (\mathcal{J}_{\mathrm{ee}}\!\times\!\mathcal{J}_{\mathrm{c}})
 \cup
 (\mathcal{J}_{\mathrm{c}}\!\times\!\mathcal{J}_{\mathrm{ee}}),
\end{equation}
where the first and second elements index participants $A$ and $B$, respectively. Thus, every optimized contact involves at least one limb end effector. For each $(i,j)\in\mathcal{P}_{\mathrm{ee}}$, we compute
\begin{equation}
 d_{t,i,j}^{\mathrm{src}}
 =
 \left\|
 j_{t,i}^{A,\mathrm{src}}
 -j_{t,j}^{B,\mathrm{src}}
 \right\|_2.
\end{equation}
A candidate pair is considered to be in contact whenever its distance is below the source detection threshold $\tau_{\mathrm{src}}$:
\begin{equation}
 \mathcal{C}_t^{\mathrm{src}}
 =
 \left\{(i,j)\in\mathcal{P}_{\mathrm{ee}}
 \;\middle|\;
 d_{t,i,j}^{\mathrm{src}}<\tau_{\mathrm{src}}
 \right\}.
 \label{eq:source-contact-retarget}
\end{equation}
Consecutive detections of the same joint pair are grouped into contact episodes, and isolated detections shorter than a minimum duration are discarded. Restricting the candidate set is important for interactions such as hugging, in which many upper-body joints may be spatially close but should not all be pulled together. We retain the selected joint identities throughout retargeting using the source-to-target joint correspondence provided to GMR. Consequently, a selected hand or foot contact must be restored using the corresponding end-effector joint and cannot be replaced by bringing a nearby wrist, forearm, ankle, or lower-leg joint close to the other participant.

\paragraph{Contact-aware Local Refinement.}
Starting from the GMR solution, we jointly refine the two target configurations using forward kinematics. Let $\phi(i)$ denote the target joint corresponding to source joint $i$, and let $J_{\phi(i)}^p(q_t^p)$ be its world position under target configuration $q_t^p$. For each retained source contact $(i,j)$, the corresponding target-joint distance is
\begin{equation}
 d_{t,i,j}^{\mathrm{tar}}
 =
 \left\|
 J_{\phi(i)}^A(q_t^A)
 -J_{\phi(j)}^B(q_t^B)
 \right\|_2.
\end{equation}
We use a target tolerance $\tau_{\mathrm{tar}}$, separate from the more permissive source detection threshold $\tau_{\mathrm{src}}$, and minimize
\begin{equation}
 \mathcal{L}_{\mathrm{contact}}
 =
 \sum_{t}
 \sum_{(i,j)\in\mathcal{C}_t^{\mathrm{src}}}
 \left[
 \max\!\left(
 d_{t,i,j}^{\mathrm{tar}}-\tau_{\mathrm{tar}},0
 \right)
 \right]^2.
 \label{eq:retarget-contact-loss}
\end{equation}
Because $\mathcal{C}_t^{\mathrm{src}}$ contains only end-effector-involving pairs, the refinement targets hand and foot contacts rather than indiscriminately attracting all nearby body joints. The joint correspondence $\phi$ also prevents a selected contact from being satisfied by switching the optimized target from the designated end effector to a nearby joint. The one-sided loss restores a missing contact but applies no additional attraction after the designated joints are sufficiently close.

For efficiency and locality, we optimize only temporally padded windows around the detected contact episodes. For an end-effector--body contact, we update the kinematic chain of the contacting limb while the counterpart body joint remains on its GMR trajectory. This corresponds primarily to the shoulder--elbow--wrist--hand chain for a hand contact and the hip--knee--ankle--foot chain for a foot contact. When both members of a pair are end effectors, we update the corresponding limb chains of both participants. All unrelated joint trajectories remain fixed to the GMR result. Let $\mathcal{Q}_{\Omega}$ collect these selected joint variables within an optimized window $\Omega$. We regularize them toward the GMR solution and preserve its temporal changes:
\begin{align}
 \mathcal{L}_{\mathrm{pose}}
 &=
 \sum_{p\in\{A,B\}}
 \sum_{t\in\Omega}
 \left\|q_t^p-\widetilde{q}_t^p\right\|_{W_q}^{2},\\
 \mathcal{L}_{\mathrm{temp}}
 &=
 \sum_{p\in\{A,B\}}
 \sum_{t,t-1\in\Omega}
 \left\|
 (q_t^p-q_{t-1}^p)
 -(\widetilde{q}_t^p-\widetilde{q}_{t-1}^p)
 \right\|_2^2.
\end{align}
The root and joints outside the active limb chains receive substantially larger entries in $W_q$, preventing the optimizer from recovering a local end-effector contact by translating the full character or disturbing unrelated parts of its support configuration. The configurations at the boundaries of each padded window are fixed to the GMR trajectory. The final refinement solves
\begin{equation}
 \min_{\mathcal{Q}_{\Omega}}
 \lambda_{\mathrm{contact}}\mathcal{L}_{\mathrm{contact}}
 +\lambda_{\mathrm{pose}}\mathcal{L}_{\mathrm{pose}}
 +\lambda_{\mathrm{temp}}\mathcal{L}_{\mathrm{temp}},
 \label{eq:contact-aware-retarget}
\end{equation}
subject to the target humanoid's joint limits. Because the source contact pairs are detected only once, each optimization step evaluates only the forward-kinematic positions of a small set of target joints. We use the refined trajectories as the unified references for training and evaluating the interaction tracking policy.

\section{Tracking Policy}
\label{app:tracking}
\paragraph{Humanoid Control and Observation.}
Following PHC~\citep{luo2023perpetual}, we represent each participant by a 24-body neutral-SMPL humanoid with 23 actuated non-root joints. The policy operates at 30~Hz and outputs a 69-dimensional action specifying three-dimensional target rotations for these joints; the simulator runs at 60~Hz and applies the targets through PD controllers. The normalized policy output $a_t^p$ for participant $p$ is converted to a generalized-position target as
\begin{equation}
 q_{t,\mathrm{PD}}^p
 =q_{\mathrm{offset}}+q_{\mathrm{scale}}\odot a_t^p,
\end{equation}
where the offset and scale are determined by the joint limits.

For each participant, the observation contains both self and partner information. The current-state descriptor includes root height and the heading-local body positions, six-dimensional rotations, linear velocities, and angular velocities. The reference descriptor encodes the corresponding differences from the next reference frame, together with its body poses in the same local representation. Denoting the other participant by $\bar p$, the policy is conditioned on
\begin{equation}
 o_t^p=
 \left[o_{t,\mathrm{self}}^p,
       o_{t,\mathrm{ref}}^p,
       o_{t,\mathrm{self}}^{\bar p},
       o_{t,\mathrm{ref}}^{\bar p}\right].
\end{equation}
Conditioning on the complete partner observation allows the tracker to respond to the partner's realized state rather than independently tracking the two reference motions.

\paragraph{Cross-attention Policy.}
The self and partner descriptors are processed by separate MLP encoders, producing features $f_t^p,f_t^{\bar p}\in\mathbb{R}^{512}$. Each feature is reshaped into four 128-dimensional tokens. We use the self tokens as queries and the partner tokens as keys and values in a four-head cross-attention layer:
\begin{equation}
 h_t^p
 =
 \operatorname{LN}\!\left(
 f_t^p+
 \operatorname{vec}\!\left[
 \operatorname{MHA}
 \left(T_t^p,T_t^{\bar p},T_t^{\bar p}\right)
 \right]
 \right).
 \label{eq:tracker-cross-attention}
\end{equation}
The residual feature $h_t^p$ is passed to an MLP policy head that predicts the action distribution, and we use its mean action for deterministic rollout. The critic uses the same interaction-aware encoding structure. In contrast to independently controlled trackers, this design lets each humanoid modulate its action according to the current and desired motion of its partner.

\paragraph{Interaction-aware Reward.}
The base imitation objective follows PHC and measures errors in body positions, rotations, linear velocities, and angular velocities. Writing these errors as $E_p,E_R,E_v,$ and $E_\omega$, respectively, the imitation reward is
\begin{equation}
 r_{\mathrm{imit}}
 =0.5e^{-100E_p}
 +0.3e^{-10E_R}
 +0.1e^{-0.1E_v}
 +0.1e^{-0.1E_\omega}.
 \label{eq:tracker-imitation-reward}
\end{equation}
We further introduce an interaction-graph reward~\citep{zhang2023simulation,li2026interagent} to preserve the spatial and dynamic relationships between the two humanoids. For every cross-person body pair $(i,j)$, let
\begin{equation}
 d_{ij}=\left\|p_i^A-p_j^B\right\|_2,
 \qquad
 u_{ij}=v_j^B-v_i^A
\end{equation}
be its distance and relative velocity, with $\bar d_{ij}$ and $\bar u_{ij}$ denoting the corresponding reference quantities. Using proximity-aware edge weights $\alpha_{ij}$, we compute
\begin{align}
 E_{\mathrm{G}}^{p}
 &=\sum_{i,j}\alpha_{ij}(d_{ij}-\bar d_{ij})^2,\\
 E_{\mathrm{G}}^{v}
 &=\sum_{i,j}\alpha_{ij}
 \left\|u_{ij}-\bar u_{ij}\right\|_2^2,\\
 r_{\mathrm{G}}
 &=0.8e^{-20E_{\mathrm{G}}^{p}}
 +0.2e^{-0.5E_{\mathrm{G}}^{v}}.
 \label{eq:tracker-graph-reward}
\end{align}
The task reward combines individual motion tracking and interaction preservation as
\begin{equation}
 r_{\mathrm{task}}
 =0.6\left(r_{\mathrm{imit}}+r_{\mathrm{power}}\right)
 +0.4r_{\mathrm{G}},
\end{equation}
where $r_{\mathrm{power}}$ penalizes excessive joint power. Inter-person collision is enabled during training and execution. We optimize the actor and critic using PPO with the adversarial motion-prior objective retained from PHC.

\section{Text-Motion Evaluators}
\label{app:tmr}
We train separate text-motion evaluators InterCLIP and InterCLIP-Phys for the digital motion domain and the physically executed humanoid domain, respectively. InterCLIP evaluates generated digital motion before execution and is trained on the InterHuman and Inter-X training sets using source-balanced sampling. Following the standard interaction representation, it encodes joint positions, frame displacements, and local joint rotations for both participants, while excluding the foot-contact indicators. InterCLIP-Phys evaluates the tracker-executed trajectories. The metrics related to text-motion consistency in Tab.~\ref{tab:interagent} are computed by it. It encodes the positions, orientations, linear velocities, and angular velocities of the 24 rigid bodies of both humanoids, expressed in an actor-centric heading-local frame. Both evaluators adopt the TMR-style retrieval architecture~\citep{petrovich2023tmr}. A frozen CLIP ViT-L/14 text tower followed by a two-layer trainable adapter encodes captions, while an eight-layer temporal Transformer encodes motion sequences. The two branches produce $\ell_2$-normalized 512-dimensional embeddings. Let $z_i^m$ denote the embedding of motion $i$, $z_j^t$ the embedding of caption $j$, and $P_i$ the captions associated with motion $i$. Their similarity is
\begin{equation}
 s_{ij}=\frac{(z_i^m)^\top z_j^t}{\tau},
\end{equation}
where $\tau$ is a temperature hyperparameter that controls the concentration of the text--motion similarity distribution.
We optimize the symmetric multi-positive contrastive objective
\begin{align}
 \mathcal{L}_{m\rightarrow t}
 &=-\frac{1}{B}\sum_{i=1}^{B}\frac{1}{|P_i|}
 \sum_{j\in P_i}
 \log\frac{\exp(s_{ij})}{\sum_{k\in V_i}\exp(s_{ik})},\\
 \mathcal{L}_{\mathrm{eval}}
 &=\frac{1}{2}\left(
 \mathcal{L}_{m\rightarrow t}+\mathcal{L}_{t\rightarrow m}
 \right),
\end{align}
where $V_i$ is the set of valid caption candidates and $\mathcal{L}_{t\rightarrow m}$ is defined analogously. All captions describing the same motion are treated as positives. Identical normalized captions associated with different motion identities are removed from the negative set to avoid explicit false negatives.

\paragraph{Physical Evaluator Data.}
\label{app:interclip-phys}
Following the evaluation protocol of InterAgent~\citep{li2026interagent}, InterCLIP-Phys is trained exclusively on physical executions of InterHuman training motions. We remove duplicated two-person motions across splits, including participant-swapped duplicates, and execute the remaining reference interactions with a frozen two-person tracking policy. To ensure a fair comparison, this policy is distinct from the pretrained tracker used by DIGHT, preventing the evaluator construction from favoring our method. We retain complete, finite, non-falling rollouts with mean body tracking distance below $0.25$~m. During training, we sample motion identities uniformly, select one accepted execution, apply temporal cropping, and randomly exchange the two participants.

\end{document}